\documentclass{article}

\PassOptionsToPackage{round, comma, sort&compress, authoryear}{natbib}
\usepackage[preprint]{neurips_2025}
\usepackage[T1]{fontenc}
\usepackage[utf8]{inputenc}
\usepackage{microtype}
\usepackage{textcomp}

\PassOptionsToPackage{hyphens}{url}
\usepackage{url}
\usepackage{amsmath}
\usepackage{amssymb}

\usepackage{booktabs}
\usepackage{array}
\usepackage{longtable}
\usepackage{calc}              
\providecommand{\real}[1]{#1}
\usepackage{enumitem}

\usepackage[font=small,labelfont=bf]{caption}

\usepackage{graphicx}
\graphicspath{{figures/}}

\usepackage{xcolor}

\usepackage[colorlinks=true,
            linkcolor=blue!50!black,
            citecolor=blue!50!black,
            urlcolor=blue!50!black]{hyperref}
\usepackage[capitalize, noabbrev]{cleveref}

\title{When Probing Accuracy Saturates, Fragility Resolves:\\
  A Complementary Metric for LLM Pre-Training Analysis}

\author{%
  Orion Reblitz-Richardson\thanks{%
    Distiller Labs. Correspondence to Distiller Labs
    \textless\texttt{orion@orionr.com}\textgreater.}
}

\date{August 2026}

\providecommand{\tightlist}{%
  \setlength{\itemsep}{0pt}\setlength{\parskip}{0pt}}

\usepackage{fancyvrb}
\DefineVerbatimEnvironment{Highlighting}{Verbatim}{commandchars=\\\{\}}
\newenvironment{Shaded}{\begin{quote}}{\end{quote}}

\newcommand{\CommentTok}[1]{\textit{#1}}

\newcommand{\NormalTok}[1]{#1}

\newcommand{\ExtensionTok}[1]{#1}

\newcommand{\AttributeTok}[1]{#1}

\begin{document}

\maketitle

\begin{abstract}
Standard linear probing declares a property ``encoded'' when a
classifier on hidden states achieves high accuracy. The protocol works
well on a snapshot but breaks across pre-training, with probe accuracy
saturating within the first few thousand steps, leaving most of
training invisible to the instrument.

We introduce \emph{fragility}, a complementary per-layer metric
defined as the activation-noise level at which probe accuracy
collapses. Fragility is sensitive to both the margin of separability
and the redundancy of representation, both of which keep evolving long
after accuracy plateaus.

Applied to open-checkpoint language models, fragility recovers
structure that accuracy alone cannot see. Moralized representations,
our interest, emerge in stages, with high-accuracy detection of morally
loaded words first and compositional encoding later. Because probe
accuracy on its own tracks how lexically separable a dataset is, we
establish the compositional encoding directly, by showing it transfers
across construction types that share no contrast tokens.
\emph{Where probing accuracy returns a flat answer, fragility returns a
structured one.}

Raw critical noise also shows a layer-depth gradient, but we identify
and correct a confound that, to our knowledge, has gone undocumented in
noise-injection probing. That is, deeper layers carry larger
activations, so a fixed absolute noise makes them look more robust than
they are. Rescaling per layer removes most of the gradient, while
within-layer comparisons stay valid and still separate
identical-accuracy corpora by fragility. \emph{We recommend
RMS-normalized critical noise for cross-layer claims and raw critical
noise for within-layer comparisons.}
\end{abstract}

\section{Introduction}\label{introduction}

A standard interpretability protocol works as follows: given a
representation of interest from a large language model, train a
linear classifier to predict the property from frozen hidden states,
report classifier accuracy at each layer, and declare the property
``linearly encoded'' wherever accuracy is high. The protocol is well-
established \citep{alain2017probes,belinkov2022probing} and works well for
asking \emph{whether} a model represents a property at a given snapshot.

It does not work well for asking \emph{how a representation evolves
during pre-training}. We demonstrate the failure mode concretely. On
the OLMo-2 1B early-training trajectory \citep{olmo2_2025},
37 model checkpoints densely sampled at 1K-step intervals across
steps 0-36K (\textasciitilde76B tokens), a binary linear probe trained on a
240-pair moral / neutral minimal-pair dataset reaches \textasciitilde95\% mean
accuracy across all 16 transformer layers by step 4K. For the
remaining \textasciitilde33K training steps (\textasciitilde95\% of the trajectory we have data
for), the standard probing instrument returns essentially the same
number; whatever continued representational change the model
undergoes through that period is invisible to it.

This paper makes a methodological contribution that takes the
saturation problem as a fixed feature of probing accuracy and adds a
complementary metric to recover the missing resolution:
\textbf{\emph{fragility}}, defined as the activation-noise level at which probe
accuracy drops below a threshold. Formally, for layer $\ell$ with trained
probe \(f_\ell\), standard probing reports test-set accuracy
\(A(f_\ell)\). We define the \textbf{critical noise}
\(\sigma^*_\ell\) as the smallest noise scale at which accuracy under
Gaussian perturbation drops below a fragility threshold \(\tau\):

\[\sigma^*_\ell = \min\bigl\{\,\sigma \in \mathcal{S} : A\!\bigl(f_\ell,\; h_\ell + \varepsilon\bigr) < \tau,\quad \varepsilon \sim \mathcal{N}(0,\,\sigma^2 I)\,\bigr\}\]

where \(\mathcal{S} = \{0.1,\, 0.3,\, 1.0,\, 3.0,\, 10.0\}\) and
\(\tau = 0.6\) (if no \(\sigma\) in \(\mathcal{S}\) brings accuracy below
\(\tau\), \(\sigma^*_\ell = \max(\mathcal{S}) = 10.0\)). A low
\(\sigma^*_\ell\) means the representation at
layer $\ell$ is \textbf{fragile}: probe accuracy collapses under small noise.
A high \(\sigma^*_\ell\) means the encoding is \textbf{robust}: the
distinction is encoded with wide margin and/or redundancy. Fragility
is a per-layer measurement applied to the same trained probe used for
the accuracy curve, and it is sensitive to both the \emph{margin} of
separability and the \emph{redundancy} of representation, both of which
keep evolving through training even after accuracy has plateaued
(it does not separately identify their contributions; see \Cref{why-fragility-succeeds-where-accuracy-saturates}). We use fragility
to map structural representational change that probing accuracy alone
cannot see, and to establish three findings on the OLMo-2 1B and
OLMo-3 7B open-checkpoint family that together earn the
methodological claim its keep:

\textbf{Finding 1: Moralized representations emerge in stages, with
high-accuracy detection of morally loaded words first and
compositional encoding later.} Through diverse
probes, we find moral encoding is acquired progressively, in order
of semantic complexity. A standard moral
probe (single morally-loaded lexeme swap) onsets at step 1K. A
\emph{compositional} moral probe (pairs that hold the action verb
constant and vary only individually-mild tokens whose moral status
flips in context (``protect'' / ``humiliate'', ``hungry'' / ``wealthy'',
``innocent'' / ``guilty'') onsets at step 5K under 4-seed averaging
(per-seed range 4K-7K), between sentiment (2K) and syntax (6K).
The standard probe's step-1K onset measures how quickly moralized
vocabulary becomes linearly separable, not how quickly moral
valence is encoded compositionally; onset timing tracks lexical
accessibility, so we establish the compositional encoding itself by
held-out transfer across construction types that share no contrast
tokens (\Cref{emergence-ordering}), not by onset order.

\textbf{Finding 2: Raw critical noise is confounded by activation scale;
we isolate the confound and establish when the metric is valid.}
In raw units the standard probe shows a striking layer-depth
robustness gradient that keeps evolving long after accuracy plateaus
(late layers tolerate far more noise than early ones, and mean
critical noise moves through step 36K). But per-layer activation RMS
grows \textasciitilde8$\times$ from early to late layers, so a fixed raw $\sigma$ is a much larger
relative perturbation early than late. An RMS-normalized control
(\S4.4) shows the cross-layer gradient and its post-saturation evolution
are largely this scale effect: the late/early ratio collapses from
\textasciitilde7--15$\times$ to \textasciitilde2$\times$ and the post-saturation decline flattens, for both the
standard and compositional probes. The confound is strictly cross-layer
and leaves within-layer comparisons intact. The practical contribution
is the resulting recommendation: raw $\sigma$* for within-layer comparisons
(where scale is fixed by construction) and RMS-normalized $\sigma$* for
cross-layer claims.

\textbf{Finding 3: Data curation reshapes probe robustness without changing
probe accuracy.} LoRA fine-tuning on three matched corpora (narrative-
moral, declarative-moral, general non-moral control) produces
identical probing accuracy across conditions (final peak 0.740 /
0.750 / 0.750) but distinct fragility profiles. Declarative moral
training (``Stealing is wrong'' repeated) produces fragility dips
at 10 of 16 layers (mean critical noise 5.63) versus 6-7 fragile
layers for natural-text conditions (mean 6.94 / 7.38). Accuracy says ``no signal'';
fragility says ``declarative training creates broadly fragile
representations.''

The unifying claim is methodological, not moral-domain-specific:
\textbf{in every comparison we test, where probing accuracy returns a flat
answer, fragility returns a structured one.} \Cref{related-work} places the work against
related literatures; \Cref{methodology} details the four minimal-pair datasets,
linear probing, and the fragility test; \Cref{results} reports results; \Cref{discussion}
discusses the phase-transition-vs-gradual-emergence taxonomy implied
by Finding 1, the geometric reasons fragility succeeds where
accuracy saturates, and limitations; \Cref{conclusion} concludes.

\section{Related work}\label{related-work}

\textbf{Linear probing.} \citet{alain2017probes} established linear
probing as a layer-wise diagnostic for what intermediate
representations contain; \citet{belinkov2022probing} surveys the methodology's
promise and known limitations. Subsequent work has formalized
accuracy-based probing's structural shortcomings via control tasks
\citep{hewitt2019control}, information-theoretic probing \citep{pimentel2020information}, and minimum-description-length analyses \citep{voita2020mdl}. Our methodological contribution extends the saturation
concern: where the probing literature treats ceiling effects as a
threat to validity, we treat them as a fixed feature of the
instrument and add a complementary metric (fragility) that continues
to resolve representational change after accuracy plateaus.

\textbf{Activation perturbation.} \citet{borras2022walking} proposed ``Walking
Noise,'' injecting additive Gaussian noise at individual layers and
defining a per-layer midpoint noise level, the closest prior
concept to our per-layer critical noise. However, Walking Noise
measures end-to-end \emph{task accuracy} degradation, not probe accuracy
on specific concepts, and evaluates a single trained model rather
than tracking across training checkpoints. APEX \citep{ren2026apex}
injects Gaussian noise into hidden activations and defines ``escape
noise'' (the noise scale at which output becomes independent of
input), analogous to our critical noise; but APEX measures
model-output distribution changes, not concept-specific probe
accuracy. Our method combines per-layer noise robustness with
concept-specific probing \emph{across} training checkpoints, a
combination that neither approach attempts.

\textbf{Probing across training.} \citet{qian2024trustworthiness} applied linear probes
to 360 pre-training checkpoints (LLM360 Amber 7B) for five
trustworthiness dimensions, observing a fitting-and-compression
pattern. Their work establishes the ``probing across checkpoints''
approach but tracks only probe \emph{accuracy}, which is exactly what
we show saturates early and stops returning information. Our
fragility metric recovers the dynamics that continue after the
point where their approach loses resolution.

\textbf{Causal tracing.} \citet{meng2022rome} introduced ROME and the
causal-tracing methodology, which distinguishes layers that \emph{encode}
information from layers that \emph{causally use} it. As an uncontrolled
exploratory observation, our 7B causal-probing analysis (Appendix B)
finds the probing peak and the causal peak for moral information fall
in different parts of the network. We report it as a direction for
future work, not as a validated result or as supporting evidence for
the body's thesis: the peak-layer argmax is unstable across probe
seeds and the tracer has no specificity control (Appendix B).

\textbf{Phase transitions.} \citet{power2022grokking} documented ``grokking,''
sudden phase transitions from memorization to generalization, with
subsequent mechanistic work attributing such transitions to discrete
circuit formation \citep{olsson2022induction,nanda2023progress}. Our \Cref{emergence-ordering}
finding that semantic minimal-pair tasks (standard moral, sentiment)
emerge as sharp phase transitions while compositional and structural
tasks emerge gradually maps the grokking phenomenon against a
lexical-vs-compositional dichotomy \emph{within a single training run},
to our knowledge a novel framing.

\textbf{Moral Foundations Theory.} The standard moral dataset organizes
content across \citeauthor{haidt2012righteous}'s \citeyearpar{haidt2012righteous} and \citeauthor{graham2013mft}'s \citeyearpar{graham2013mft} six MFT
foundations (care/harm, fairness/cheating, loyalty/betrayal,
authority/subversion, sanctity/degradation, liberty/oppression). MFT
is used as a \emph{construction} heuristic for balanced coverage, not as
a cognitive claim about how language models represent morality. The
compositional dataset (\Cref{compositional-moral-probing-dataset}) categorizes by construction pattern
(motive / target / consequence / role) instead.

\textbf{OLMo.} \citet{groeneveld2024olmo} released the OLMo family with
full intermediate checkpoint releases (the infrastructure that
makes dense-sampling trajectory analysis possible), extended in
the OLMo-2 release \citep{olmo2_2025} to substantially longer
training schedules. Our 37-checkpoint 1B early-training and
20-checkpoint 7B stage-1 trajectories rely on these open releases;
the methodology applies to any open-checkpoint family with dense
enough sampling. Pythia \citep{biderman2023pythia} offers a
complementary open-checkpoint testbed.

\textbf{Single-direction circuits.} \citet{arditi2024refusal} demonstrated
refusal behavior in instruction-tuned models is mediated by a single
representational direction. The fragility metric is the pre-training-
time analog: where single-direction-refusal asks whether \emph{post-
training} safety properties are concentrated in narrow circuits at
the final checkpoint, we ask whether moralized representations pass
through more or less brittle states \emph{during} training. Both
literatures address how concentrated vs.~distributed safety-relevant
representations are, at different points in the model lifecycle.
Representation-engineering more broadly \citep{zou2023repe} treats
interpretability as direct readout of learned representations.

\textbf{Scope.} We use moralized vocabulary as a demonstration domain.
The compositional probe (\Cref{compositional-moral-probing-dataset}) is the explicit lexical-
accessibility ablation; deeper questions (counterfactual moral
reasoning, generalization to held-out moral structures) would
require harder probes and are discussed as future work in \Cref{limitations}.

\section{Methodology}\label{methodology}

We apply linear probing classifiers on four matched minimal-pair
datasets to all 37 OLMo-2 1B early-training checkpoints, 20 OLMo-3
7B stage-1 checkpoints, and the OLMo-2 1B final checkpoint. Two
probe families: \texttt{LayerWiseMoralProbe} (per-layer accuracy) and
\texttt{MoralFragilityTest} (per-layer noise robustness). All experiments
run on a single MacBook Pro M4 Pro / MPS; code, datasets, and
per-checkpoint outputs are released with the paper.

\subsection{Standard minimal-pair datasets}\label{standard-minimal-pair-datasets}

Three single-token-swap datasets mirror established minimal-pair
probing practice \citep{belinkov2022probing} by holding the syntactic skeleton
constant and swapping a single token: \textbf{moral / neutral} (240 pairs,
40 per Moral Foundations Theory category (care/harm,
fairness/cheating, loyalty/betrayal, authority/subversion,
sanctity/degradation, liberty/oppression; Haidt, 2012; Graham et
al., 2013; e.g.~``She betrayed the woman'' / ``She greeted the
woman''). The moral pairs are drawn from a 1,200-pair dataset
constructed per published quality guidelines with LLM-assisted
filtering for naturalness and moral neutrality of neutral-side
sentences (see \path|DATASET_GUIDELINES.md|); 240 pairs (40 per
foundation) are subsampled with a deterministic seed.
\textbf{Sentiment} (210 pairs across ten domains; positive /
negative adjective swap, e.g.~``\ldots excellent food\ldots{}'' / ``\ldots terrible
food\ldots{}''); \textbf{syntax} (210 pairs targeting local grammaticality
violations (word-order swaps, agreement, auxiliary misplacement),
e.g.~``She handed the finished report\ldots{}'' / ``She handed finished the
report\ldots{}''). All three use a length-ratio gate $\le$ 1.5 and a
deterministic seed-42 train / test split. They establish the
baseline emergence ordering in \Cref{emergence-ordering}.

\subsection{Compositional moral probing dataset}\label{compositional-moral-probing-dataset}

The standard moral probe's pairs differ in a single morally-loaded
lexeme, which a linear probe can separate from single-token
distributional statistics alone. The standard probe's step-1K onset
therefore measures \emph{how quickly moralized vocabulary becomes
linearly separable from neutral vocabulary}, not \emph{how quickly moral
valence is encoded compositionally}. To bound this lexical-accessibility
framing empirically, we construct a compositional moral minimal-pair
dataset whose pairs require multi-word integration to determine moral
valence. This dataset is a primary methodological contribution of the
paper.

\textbf{Construction.} Each pair holds the action verb constant and varies
1--2 individually-mild tokens whose moral status only resolves in the
surrounding action context. Four 50-pair categories (200 pairs total),
all hand-curated:

\begin{itemize}
\tightlist
\item
  \emph{action\_motive:} same action verb, motive determines valence.
  ``She lied to her parents to \textbf{protect} her younger brother today'' /
  ``\ldots to \textbf{humiliate}\ldots{}''
\item
  \emph{action\_target:} same action, target descriptor determines moral
  relevance. ``He gave the last loaf of bread to the \textbf{hungry} stranger
  at the door'' / ``\ldots the \textbf{wealthy} stranger\ldots{}''
\item
  \emph{action\_consequence:} same action, consequence framing determines
  valence. ``He kept the secret about the surprise to keep his sister
  \textbf{safe} today'' / ``\ldots to keep his sister \textbf{hurt} today''
\item
  \emph{role\_reversal:} same components, role / target / context
  determines valence. ``The judge accepted the gift to free the
  \textbf{innocent} prisoner from prison'' / ``\ldots the \textbf{guilty}
  prisoner\ldots{}''
\end{itemize}

Contrast tokens are individually mild: \texttt{protect}, \texttt{humiliate},
\texttt{hungry}, \texttt{wealthy}, \texttt{safe}, \texttt{hidden}, \texttt{innocent}, \texttt{guilty}. None
appear in the strong-valence blocklist of 47 tokens (\texttt{murder},
\texttt{torture}, \texttt{stole}, \texttt{rape}, \texttt{assault}, etc.) we use as a construction
gate. Each token is unmarked or only weakly valenced in isolation;
the moral signal in each pair lives in the integration of the
contrast token with the surrounding action context.

\textbf{Validation gates.} All 200 pairs pass five gates simultaneously
(\path|validate_compositional_dataset|):

\begin{enumerate}
\def\labelenumi{\arabic{enumi}.}
\tightlist
\item
  \emph{Length difference:} per-pair absolute difference $\le$ 2 alphabetic
  tokens.
\item
  \emph{Length band:} both halves in the 8-20-word band.
\item
  \emph{Content-word overlap:} per-pair $\ge$ 0.60 with stopwords removed
  (\texttt{\textbar{}shared\textbar{}\ /\ max(\textbar{}moral\textbar{},\ \textbar{}immoral\textbar{})}). This metric matches
  \path|deepsteer.datasets.validation.validate_pairs| so the 0.60
  threshold here is directly comparable to the standard moral probe
  gate.
\item
  \emph{Strong-valence blocklist:} zero tokens from the 47-word blocklist
  on either side.
\item
  \emph{No exact duplicates} of either side across the 200 pairs.
\end{enumerate}

\textbf{Operationalizing ``compositional'': transfer and lift, not a
bag-of-words gate.} A minimal pair that flips its label by swapping a
single contrast token makes that token a lexical cue \emph{by construction}:
a bag-of-words classifier separates the seen pairs no matter how mild
the token is in isolation. ``Can unigrams separate the data'' is
therefore the wrong question (they can: a pair-disjoint,
orientation-invariant unigram TF-IDF classifier reaches 0.63 overall
and 0.57-0.64 per construction, where pair-disjoint means \texttt{GroupKFold}
keyed on pair so neither half of a pair leaks across the split, and
orientation-invariant means scoring \texttt{max(acc,\ 1\ -\ acc)}). The right
question is whether the hidden-state representation encodes moral
valence in a way that generalizes \emph{beyond the specific contrast
tokens}. We operationalize compositional encoding through two tests,
both measured against this unigram TF-IDF baseline (the \emph{lexical
floor}):

\begin{enumerate}
\def\labelenumi{\arabic{enumi}.}
\tightlist
\item
  \emph{Leave-construction-out transfer.} Train the probe on three of the
  four construction categories and test on the held-out fourth, which
  shares almost no contrast tokens with the training categories. A
  probe relying on token identity cannot transfer; a
  construction-general representation can.
\item
  \emph{Lift over the lexical floor.} Within each construction, compare
  hidden-state decodability (pair-disjoint CV) to that construction's
  unigram lexical floor. The gap is signal the bag-of-words baseline
  cannot reach.
\end{enumerate}

\Cref{emergence-ordering} reports both, at the final checkpoint and across training. (The
construction iterated through \textasciitilde5 rewriting passes to satisfy the 0.60
content-overlap gate alongside the multi-word contrast requirement;
the two constraints are in genuine tension; see Appendix D.)

\textbf{Train / test split.} 160 / 40, stratified by category (40 train +
10 test per category), seed = 42. The dataset and validation gates
are deterministic, API-free, and included in the toolkit at
\path|deepsteer/datasets/compositional_moral_pairs.py|.

\subsection{Linear probing}\label{linear-probing}

Identical probing methodology across all four datasets: when we
report a 4K-step gap between standard and compositional moral
onsets (\Cref{emergence-ordering}), the only experimental variable is the dataset.

For each transformer layer $\ell$ we mean-pool hidden states across the
sequence dimension and train a binary linear classifier
(\texttt{nn.Linear(hidden\_dim,\ 1)}) to distinguish the two sides of each
minimal pair, with BCE-with-logits loss, Adam (lr = 1e-2), 50
epochs, no weight decay or early stopping; probes run in fp32 on
fp16 activation caches collected via PyTorch forward hooks on the
HuggingFace \texttt{model.layers{[}$\ell${]}} modules. We report per-layer test-set
accuracy and four summary statistics: onset layer (first $\ge$ 0.6),
peak layer, encoding depth (\texttt{onset\_layer\ /\ n\_layers}), and encoding
breadth (fraction of layers $\ge$ 0.6). The compositional probe
additionally tracks a TF-IDF content-only floor per checkpoint that
hidden-state probing must beat by $\ge$10 pp.~Implementation:
\texttt{LayerWiseMoralProbe} for standard and compositional moral (the
latter a subclass that overrides only the dataset path),
\texttt{GeneralLinearProbe} for sentiment and syntax (same training loop).

\subsection{Fragility testing}\label{fragility-testing}

Train per-layer linear probes on clean activations as in \Cref{linear-probing}; for
each layer $\ell$, add Gaussian noise N(0, $\sigma$$^2$) to the cached test-set
activations and re-evaluate the trained probe across a logarithmic
sweep $\sigma$ $\in$ \{0.1, 0.3, 1.0, 3.0, 10.0\}. The smallest $\sigma$ at which
accuracy drops below the fragility threshold (0.6, i.e.~chance + 0.1
on a binary task) is the layer's \emph{critical noise}; if no $\sigma$ in the
sweep brings the probe below threshold, critical noise is reported
as the maximum (10.0), and never-fragile layers are censored at that
maximum (not dropped) when averaging, so the mean over all layers is
\path|mean_critical_noise| as a scalar summary. The
same \texttt{MoralFragilityTest} runs against both the standard and
compositional datasets; methodology generality is established by
reuse, not reimplementation.

Two properties of this estimator bound its interpretation. First,
$\sigma$* is \textbf{floor-dominated when baseline accuracy is below $\tau$=0.6}: if
the clean probe already sits below threshold (step 0, and the
compositional probe for its first \textasciitilde2K steps), $\sigma$\emph{=0.1 by construction,
so part of the early ``fragility rise'' tracks accuracy crossing
threshold rather than a change in noise tolerance. Second, $\sigma$} is
\textbf{right-censored at the grid maximum}: layers pinned at 10.0 for the
whole trajectory could have larger true critical noise. Where
censoring is heavy (late layers throughout, and per-foundation 7B
fragility), we lift the cap with an \textbf{extended grid}
$\sigma$ $\in$ \{0.1, 0.3, 1.0, 3.0, 10.0, 30.0, 100.0\}.

\subsection{Target models and checkpoints}\label{target-models-and-checkpoints}

Three OLMo \citep{groeneveld2024olmo,olmo2_2025,olmo3_2025} base models.
\textbf{OLMo-2 1B early-training} (\path|allenai/OLMo-2-0425-1B-early-training|; Team OLMo, 2025a), 37
checkpoints at 1K-step intervals from step 0 to step 36K (\textasciitilde76B
tokens), the primary data source for \Cref{emergence-ordering} onsets and \Cref{probing-accuracy-saturates-fragility-doesnt}
fragility. \textbf{OLMo-3 7B stage 1} (\path|allenai/Olmo-3-1025-7B|; Team OLMo, 2025b):
probing on 20 stage-1 checkpoints through \textasciitilde1.4M steps (\textasciitilde10T tokens), and
fragility on 5 of those checkpoints (steps 0, 353K, 705K, 1,059K, 1,413,814);
\Cref{probing-accuracy-saturates-fragility-doesnt} 7B corroboration and Appendix B causal tracing. \textbf{OLMo-2 1B final}
(\path|allenai/OLMo-2-0425-1B|, \textasciitilde2.2T tokens), used only for the
compositional probe validation gate (\Cref{compositional-moral-probing-dataset}). All loaded in fp16 on MPS;
\textasciitilde6 hours of MPS time across the full \Cref{results} experimental record.

\subsection{Validity controls}\label{validity-controls}

Three controls standard for linear-probing studies (leave-lexeme-out
splits, paraphrase transfer, adversarial lexical swap) are described
as procedures in Appendix C. We do not run them; we rest the validity
argument instead on the compositional leave-construction-out ablation
(\Cref{compositional-moral-probing-dataset}), whose compositional encoding is established by
leave-construction-out transfer and lift over the unigram lexical floor
(\Cref{compositional-moral-probing-dataset}, \Cref{emergence-ordering}), not by the lexical floor alone, and which is a strictly
stronger ablation than those three controls combined for the relevant
question. Paraphrase transfer is the one remaining surface-variation
check we leave to future work.

\section{Results}\label{results}

\subsection{Emergence in stages: morally loaded words before compositional encoding}\label{emergence-ordering}

We train the four linear probes from \Cref{standard-minimal-pair-datasets}-\Cref{compositional-moral-probing-dataset} on hidden states from
all 37 OLMo-2 1B early-training checkpoints (steps 0-36K at 1K
intervals). The \emph{onset step} (distinct from the per-layer \emph{onset
layer} of \Cref{linear-probing}, and read at a 0.70 threshold rather than 0.60) is the
first checkpoint where mean probe accuracy across all 16 layers
reaches 0.70. \begin{figure}[t]
  \centering
  \includegraphics[width=\linewidth]{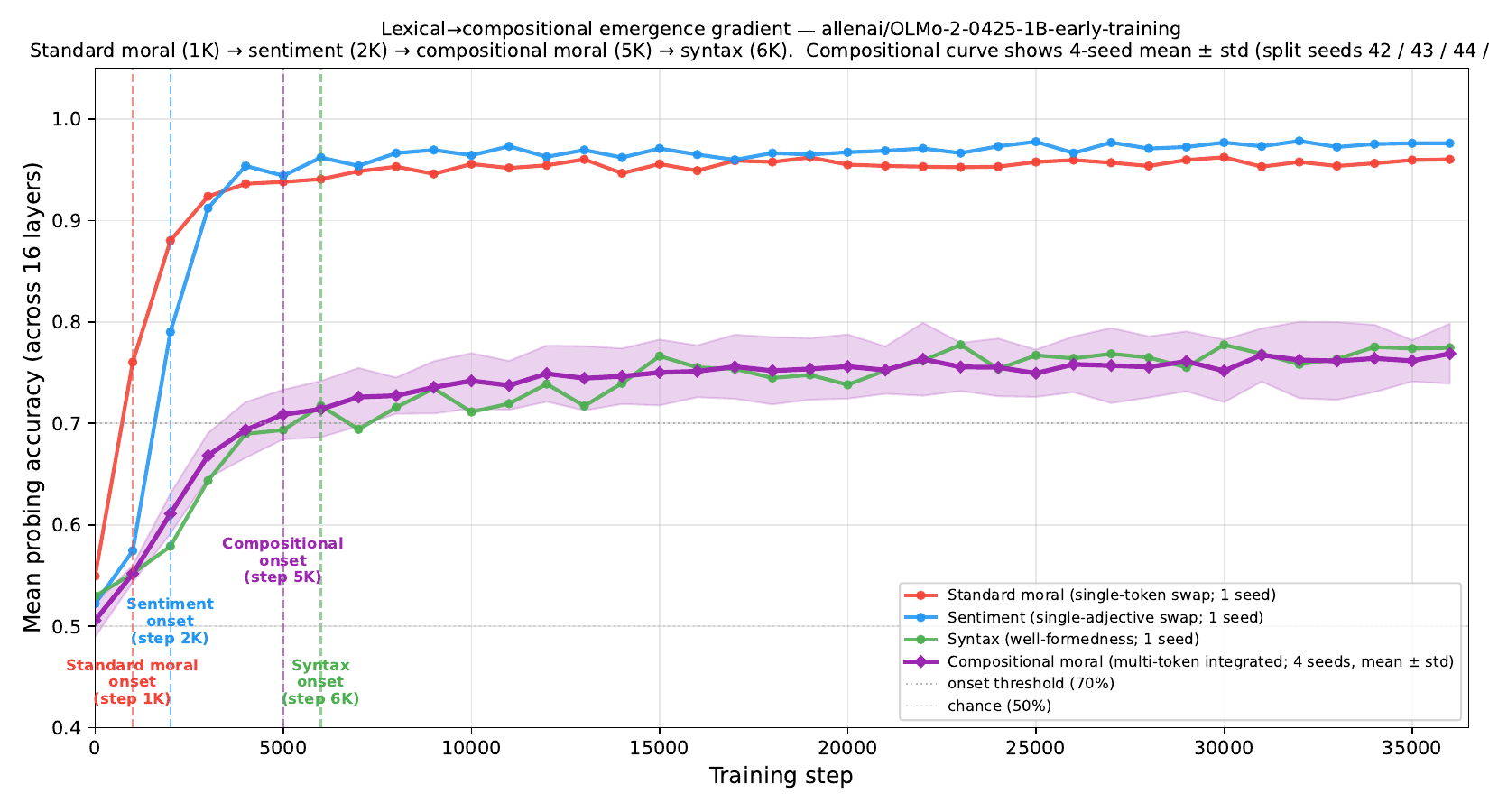}
  \caption{Staged emergence of moralized representations on OLMo-2~1B early-training. Standard moral and sentiment probes (single-token swap) plateau near~0.97; compositional moral and syntax (multi-token integration) plateau near~0.77. Compositional curve is 4-seed mean~$\pm$~std (split seeds 42/43/44/45). Onsets: standard moral~1K, sentiment~2K, compositional moral~5K (per-seed range~4K--7K), syntax~6K.}
  \label{fig:onset-overlay}
\end{figure}

\textbf{Figure~\ref{fig:onset-overlay}} plots the four mean-accuracy trajectories on a shared step axis.

{\def\LTcaptype{none} 
\begin{longtable}[]{@{}
  >{\raggedright\arraybackslash}p{(\linewidth - 8\tabcolsep) * \real{0.20}}
  >{\raggedright\arraybackslash}p{(\linewidth - 8\tabcolsep) * \real{0.27}}
  >{\raggedleft\arraybackslash}p{(\linewidth - 8\tabcolsep) * \real{0.10}}
  >{\raggedleft\arraybackslash}p{(\linewidth - 8\tabcolsep) * \real{0.16}}
  >{\raggedleft\arraybackslash}p{(\linewidth - 8\tabcolsep) * \real{0.27}}@{}}
\toprule\noalign{}
\begin{minipage}[b]{\linewidth}\raggedright
Probe
\end{minipage} & \begin{minipage}[b]{\linewidth}\raggedright
Construction
\end{minipage} & \begin{minipage}[b]{\linewidth}\raggedleft
Onset step
\end{minipage} & \begin{minipage}[b]{\linewidth}\raggedleft
Onset mean acc
\end{minipage} & \begin{minipage}[b]{\linewidth}\raggedleft
Plateau mean acc (step 36K)
\end{minipage} \\
\midrule\noalign{}
\endhead
\bottomrule\noalign{}
\endlastfoot
Standard moral & single morally-loaded lexeme swap & 1,000 & 0.760 & 0.960 \\
Sentiment & single valenced adjective swap & 2,000 & 0.790 & 0.976 \\
\textbf{Compositional moral} & \textbf{multi-token integrated swap} & \textbf{5,000} & \textbf{0.709 $\pm$ 0.025} & \textbf{0.769 $\pm$ 0.030} \\
Syntax & structural well-formedness & 6,000 & 0.717 & 0.774 \\
\end{longtable}
}

\emph{Table 1: Probe onset and plateau by construction. Compositional
moral values are 4-seed mean $\pm$ std (split seeds 42 / 43 / 44 / 45).
Per-seed compositional onsets: 4K, 4K, 7K, 7K (substantial seed
variance, with the 4-seed mean curve crossing 0.70 at step 5K). The
single-seed standard moral / sentiment / syntax curves are reported
without std bands; their seed dependence is not characterized.}

\textbf{(1) The four probes resolve into a quantitative emergence
ordering, with morally loaded words detectable first and
compositional encoding later.} The standard moral probe (single
morally-loaded lexeme swap, ``betrayed'' / ``greeted'') onsets at step
1K. The compositional moral probe (multi-token integrated swap;
contrast tokens ``protect'' / ``humiliate'', ``hungry'' / ``wealthy'' are
individually mild) onsets at step 5K under 4-seed averaging, a
4K-step lag, with per-seed onsets ranging 4K-7K and overall
trajectory always between sentiment (2K) and syntax (6K). The
standard probe's step-1K onset measures how quickly moralized
vocabulary becomes linearly separable, not how quickly moral
valence is encoded compositionally. Both findings are true; the
strongest single-token reading of the standard onset is ruled out,
while the staged reading (morally loaded words
first, compositional moral integration second, syntactic competence
last) holds. Onset ordering is a descriptive observation about
lexical accessibility, not the load-bearing evidence: the 0.709
onset sits only \textasciitilde8 pp above the 0.63 lexical floor, and onset ordering
across all four datasets tracks lexical-floor height (unigram TF-IDF
floor: standard moral 0.86, sentiment 0.80, compositional 0.63, syntax
0.59; the more lexically separable a dataset, the earlier its onset and
the higher its plateau). Onset timing on its own therefore does not
separate compositional encoding from lexical difficulty. The
load-bearing evidence that the probe recovers compositional rather
than lexical signal is the transfer-and-lift result (1b below):
held-out leave-construction-out transfer of 0.848 versus 0.598 for a
bag-of-words classifier (\Cref{compositional-moral-probing-dataset}).

\textbf{(1b) Compositional encoding is real, not lexical lookup
(final-checkpoint evidence).} At the OLMo-2 1B final checkpoint
(\textasciitilde2.2T tokens), a probe trained on three construction categories and
tested on the held-out fourth transfers at \textbf{0.848} mean (0.80-0.91
across the four held-out constructions), essentially matching its
in-distribution pair-disjoint accuracy (\textbf{0.858} @ layer 7), while a
bag-of-words classifier doing the same leave-construction-out transfer
collapses to \textbf{0.598}. Within each construction the probe decodes
+0.20 to +0.28 above the unigram lexical floor. The decisive case is
\emph{role\_reversal}, where the same components appear on both sides and the
lexical floor is lowest (0.57): hidden states still decode moral
valence at 0.85 (lift +0.28) and held-out transfer reaches 0.81. A
probe reading contrast-token identity could not do this; the model
reads moral valence from context. This is the operational content of
``compositional'' in this paper, and it is a positive result, not a thin
margin over a bag-of-words floor.

\textbf{(1c) Compositional encoding emerges in early pre-training, then
holds.} \begin{figure}[t]
  \centering
  \includegraphics[width=\linewidth]{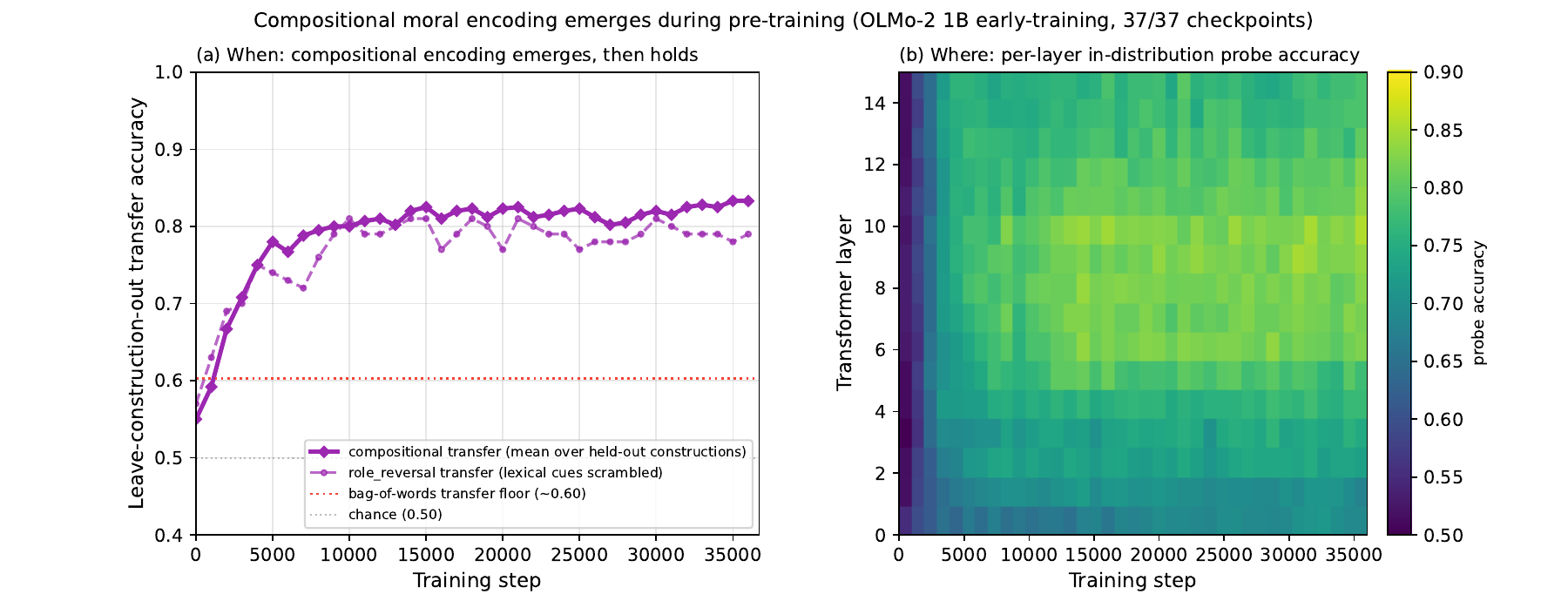}
  \caption{Compositional moral encoding emerges in early pre-training and then holds (OLMo-2~1B early-training, 37 checkpoints). (a)~When: leave-construction-out transfer accuracy rises from chance ($\sim$0.55) across steps~2K--9K, crosses the bag-of-words transfer floor ($\sim$0.60) by step~2K, and plateaus near~0.82 (lift~$\sim$+0.20 over the lexical floor) through step~36K; the role\_reversal curve (lexical cues scrambled) tracks it. (b)~Where: per-layer in-distribution probe accuracy, showing the encoding concentrating in mid-network layers~8--10 once it emerges.}
  \label{fig:compositional-emergence}
\end{figure}

\textbf{Figure~\ref{fig:compositional-emergence}} plots the transfer-and-lift analysis across all 37 early-training checkpoints. At initialization the encoding is
absent: leave-construction-out transfer sits at chance (0.55) and lift
is \textasciitilde0. It emerges over steps 2K-9K, crossing the bag-of-words transfer
floor (\textasciitilde0.60) by step 2K (0.667), passing 0.70 by step 3K, reaching
0.78 by step 5K, then plateauing by step \textasciitilde9K at transfer \textasciitilde0.82 and
lift \textasciitilde+0.20 and holding there through step 36K (0.83 / +0.22). The
role\_reversal construction, where lexical cues are scrambled by design,
follows the same curve (panel a). The encoding also localizes in
depth: once it appears, the most decodable layer settles into
mid-network (layers 8-10) and stays there for the rest of the
trajectory (panel b). Compositional moral encoding is therefore an
early-pre-training acquisition, emerging after morally loaded words
become detectable (standard-probe onset at step 1K) and consistent
with the staged ordering, that once acquired is stable across the
remaining \textasciitilde30K steps we observe. Numbers source:
\path|outputs/phase_c4_compositional/b_traj/| (per-checkpoint) and
\path|b_traj_summary.json|.

\textbf{(2) Step-like vs.~gradual emergence dichotomy.} Standard moral
and sentiment probes show sharp sigmoidal transitions (chance $\to$
plateau within one 1K-step interval at onset, then flat).
Compositional moral and syntax rise more gradually (\textasciitilde3-5K steps
across the 0.70 threshold). At 1K sampling we resolve transitions
that are step-like at this resolution; we do not claim true
discontinuity. This parallels grokking-literature observations
\citep{power2022grokking} that some capabilities emerge sharply and others
gradually; the within-run split here suggests the distinguishing
factor is whether the capability is acquirable from local lexical
statistics (sharp) or requires multi-token integration (gradual).
\Cref{semantic-vs.-structural-learning-dynamics} develops.

\textbf{(3) Plateau coincidence.} The four-curve overlay (Figure 1) makes
a structural caveat visually inescapable: probes whose signal lives
in single-token vocabulary statistics (standard moral, sentiment)
plateau at 0.96 and 0.98, while probes whose signal requires
multi-token structural or compositional integration (compositional
moral, syntax) plateau at 0.77 and 0.77. The 20-percentage-point
ceiling gap is consistent across the entire 0-36K trajectory. This
may be a probe-side property under our methodology rather than a
model property: either the 1B model encodes both compositional moral
valence and syntactic well-formedness at $\approx$0.77 (model ceiling), or
mean-pooled linear probing on 1B hidden states bottoms out at $\approx$0.77
for multi-token integration regardless of underlying representational
quality (probe ceiling). The cleanest disambiguation is repeating
\Cref{emergence-ordering} at 7B and 32B: if compositional moral rises with scale while
syntax does not, the model is the bottleneck; otherwise the probe is.
We state both readings honestly in \Cref{limitations} and refine rather than
overturn the staged-emergence finding.

\textbf{Generalization to OLMo-3 7B.} We have not yet run the compositional
probe on the OLMo-3 7B trajectory; doing so is the cleanest
disambiguation of the plateau-coincidence ambiguity and is flagged
as future work in \Cref{limitations}.

Numbers source: \path|outputs/phase_c2/c2_emergence_timing.json| (standard
moral + sentiment + syntax, 37 checkpoints) and
\path|outputs/phase_c4_compositional/c4_emergence_timing.json| (compositional
moral, 37 checkpoints; companion JSON with all four curves overlaid).
Validation source: \path|outputs/phase_c4_compositional/c4_validation.json|
(final-checkpoint validation gate on \path|allenai/OLMo-2-0425-1B|).

\subsection{Probing accuracy saturates; fragility doesn't}\label{probing-accuracy-saturates-fragility-doesnt}

\begin{figure}[t]
  \centering
  \includegraphics[width=\linewidth]{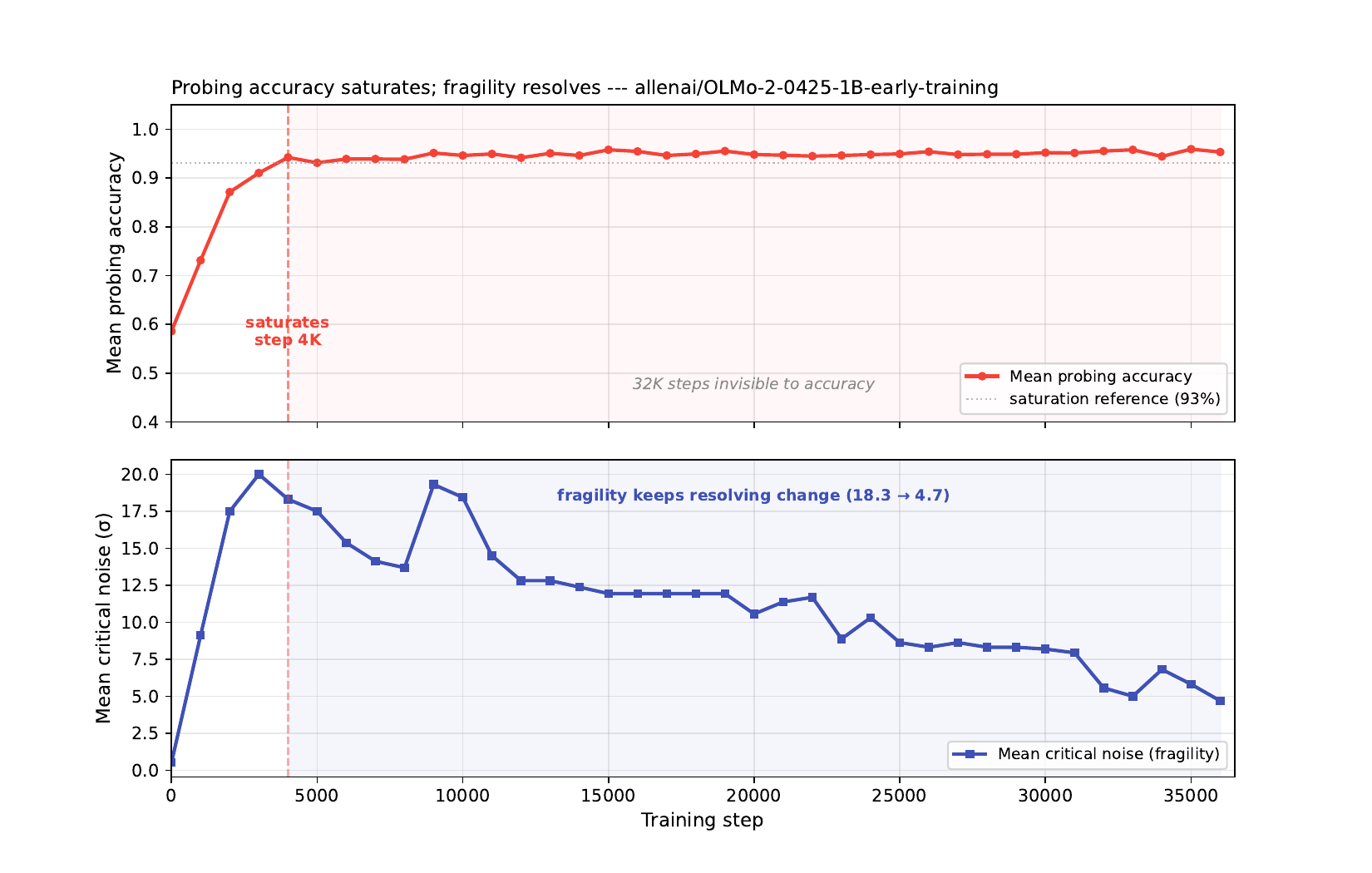}
  \caption{Probing accuracy saturates; fragility resolves. OLMo-2~1B early-training, 37 checkpoints. Top: mean probing accuracy across all 16 layers --- saturates near~0.95 by step~4K and stays flat for the remaining~32K steps. Bottom: mean critical noise --- continues evolving long after accuracy plateaus, from raw $\sigma^*\!\approx\!18.3$ at step~4K to $\approx\!4.7$ at step~36K (the values tabulated in Table~2). Note: this bottom-panel trajectory is \emph{raw} $\sigma^*$ and is largely activation scale (Sec.~4.4); the scale-free (RMS-normalized) version is flat (13.8~to~15.0).}
  \label{fig:saturation-vs-fragility}
\end{figure}

\textbf{Figure~\ref{fig:saturation-vs-fragility}} provides the central comparison for the methodological claim: a two-panel comparison on a shared step axis. Top panel:
mean probing accuracy, a sharp sigmoid from chance (\textasciitilde0.59) to a
plateau (\textasciitilde0.95) between steps 0 and 4K, then flat for the remaining
\textasciitilde33K steps. Bottom panel: mean fragility, an initial rise alongside
accuracy in the first few thousand steps, then continued movement throughout.
Top panel reaches a ceiling and stops; bottom panel keeps moving for
the entire remaining 90 \% of training. The fragility numbers below are
seed-averaged (10 noise draws per cell) on the extended noise grid
\(\sigma \in \{0.1, 0.3, 1, 3, 10, 30, 100\}\), with never-fragile
layers censored at the grid maximum (\Cref{fragility-testing}).

\textbf{OLMo-2 1B, 37 checkpoints, dense sampling.}

{\def\LTcaptype{none} 
\begin{longtable}[]{@{}
  >{\raggedleft\arraybackslash}p{(\linewidth - 10\tabcolsep) * \real{0.1667}}
  >{\raggedleft\arraybackslash}p{(\linewidth - 10\tabcolsep) * \real{0.1667}}
  >{\raggedleft\arraybackslash}p{(\linewidth - 10\tabcolsep) * \real{0.1667}}
  >{\raggedleft\arraybackslash}p{(\linewidth - 10\tabcolsep) * \real{0.1667}}
  >{\raggedleft\arraybackslash}p{(\linewidth - 10\tabcolsep) * \real{0.1667}}
  >{\raggedleft\arraybackslash}p{(\linewidth - 10\tabcolsep) * \real{0.1667}}@{}}
\toprule\noalign{}
\begin{minipage}[b]{\linewidth}\raggedleft
Step
\end{minipage} & \begin{minipage}[b]{\linewidth}\raggedleft
Mean acc
\end{minipage} & \begin{minipage}[b]{\linewidth}\raggedleft
Mean critical noise
\end{minipage} & \begin{minipage}[b]{\linewidth}\raggedleft
Late-layer crit
\end{minipage} & \begin{minipage}[b]{\linewidth}\raggedleft
Mid-layer crit
\end{minipage} & \begin{minipage}[b]{\linewidth}\raggedleft
Early-layer crit
\end{minipage} \\
\midrule\noalign{}
\endhead
\bottomrule\noalign{}
\endlastfoot
0 & 0.586 & 0.52 & 0.1 & 0.1 & 1.4 \\
1,000 & 0.731 & 9.12 & 10.0 & 10.0 & 7.2 \\
4,000 & 0.942 & 18.31 & 30.0 & 16.7 & 8.6 \\
10,000 & 0.946 & 18.44 & 44.0 & 10.0 & 3.0 \\
15,000 & 0.958 & 11.94 & 26.0 & 7.7 & 3.0 \\
20,000 & 0.948 & 10.56 & 22.0 & 7.7 & 2.6 \\
36,000 & 0.953 & 4.69 & 10.0 & 3.0 & 1.4 \\
\end{longtable}
}

\emph{Table 2: Standard moral probe, seed-averaged on the extended grid.
Accuracy plateaus by step 4K; raw critical noise keeps moving through
step 36K (rising to a peak near step 4--10K, then declining). The
extended grid lifts the late-layer band off the old \(\sigma=10\) cap
(late reaches 30--44 mid-trajectory). These are raw-unit values; \S4.4
shows that the steep late-vs-early gradient and the post-peak decline
are largely an activation-scale effect, and reports what survives
scale normalization. Read as-is, the raw numbers measure practical
perturbation sensitivity: absolute noise tolerated before the probe
degrades.}

\begin{figure}[t]
  \centering
  \includegraphics[width=\linewidth]{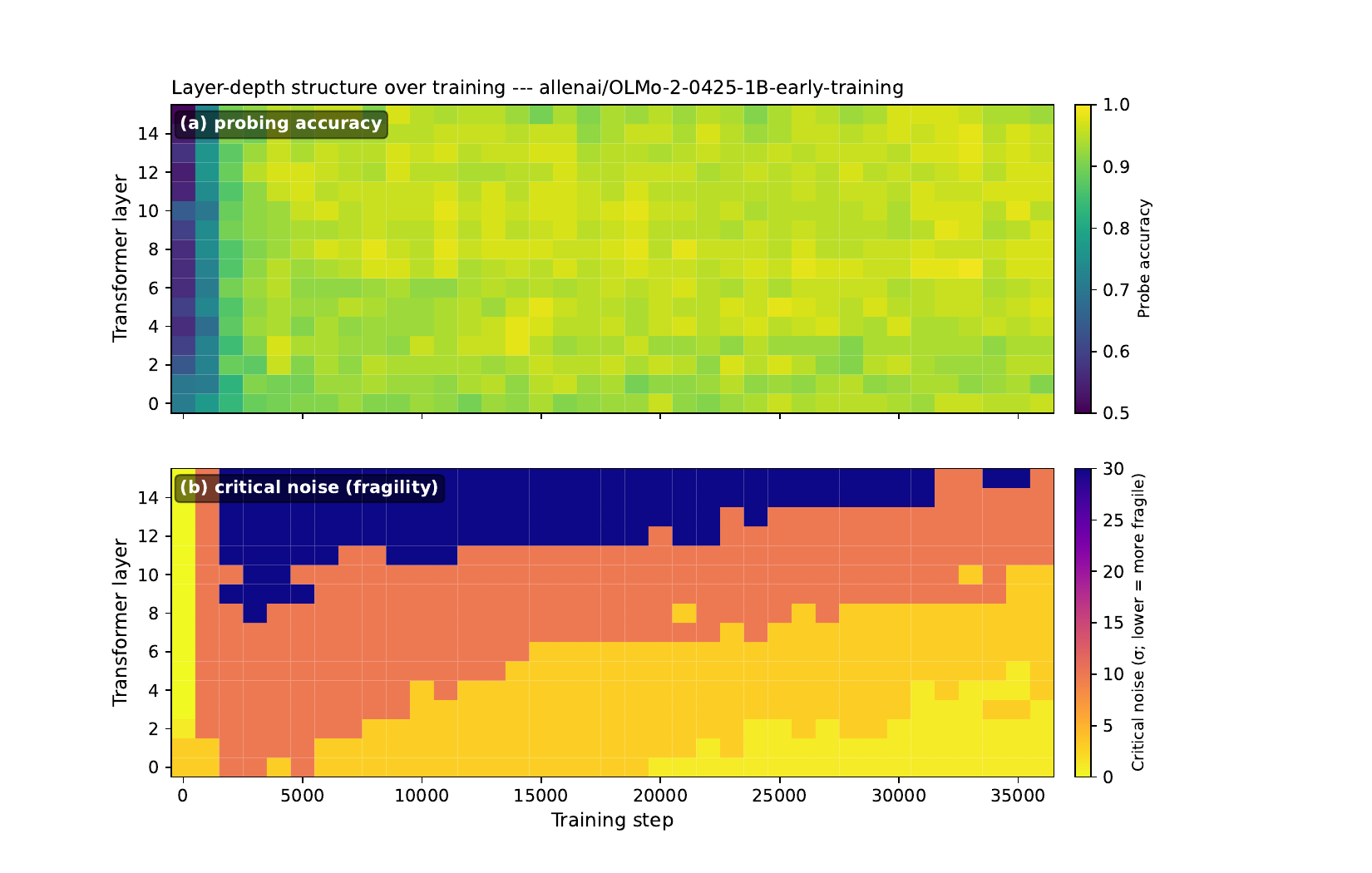}
  \caption{Layer-depth structure over training (OLMo-2~1B early-training). (a)~Probing accuracy: uniformly high across layers after step~4K. (b)~Critical noise: a layer-depth gradient develops, with late layers holding maximum noise tolerance while early layers grow progressively more brittle. Same data, same model; structure visible only under the fragility metric.}
  \label{fig:layer-depth-heatmaps}
\end{figure}

\textbf{Figure~\ref{fig:layer-depth-heatmaps}} shows the same trajectory as two stacked layer-depth heatmaps: probing accuracy (uniformly green after step 4K, no
remaining structure to resolve) above raw critical noise (a band that
deepens with layer index). The visible vertical gradient is real in
raw units but, as \S4.4 establishes, tracks the \({\sim}8\times\) growth
in activation scale from early to late layers more than probe
robustness per se. Same data; different metric; different visible
structure.

The pattern reproduces at OLMo-3 7B (5 sparse checkpoints, cap-at-max
re-aggregation of the original 5-level sweep; not seed-averaged on the
extended grid like the 1B above): mean critical noise rises
2.91 $\to$ 5.29 between steps 0 and 353K, then climbs to 6.19 by
step 1.4M. The layer-depth gradient at the final checkpoint is
late 10.0 / mid 6.2 / early 2.0 --- the same depth ordering as the 1B
and, per \S4.4, subject to the same activation-scale caveat (early
layers are smaller-norm, so raw \(\sigma\) degrades them more). The 1B
trajectory is the headline because dense 1K-step sampling resolves the
saturation step (\textasciitilde4K) and the gradient's emergence rate.

\textbf{Compositional probe fragility evolution (4-seed replication; the
methodological claim generalizes beyond the standard probe).} We
ran \texttt{MoralFragilityTest} (\Cref{fragility-testing}) on the compositional dataset across
all 37 OLMo-2 1B early-training checkpoints with four split seeds
(42, 43, 44, 45), the original seed-42 trajectory plus a three-seed
replication \textasciitilde50 min on the same MacBook Pro M4 Pro / MPS. \textbf{Table 3}
gives the 4-seed mean $\pm$ std at the diagnostic checkpoints; the
4-seed accuracy band on \textbf{Figure~\ref{fig:onset-overlay}} carries the matching probing-
side trajectory.

{\def\LTcaptype{none} 
\begin{longtable}[]{@{}rrl@{}}
\toprule\noalign{}
Step & Compositional mean critical noise (4-seed mean $\pm$ std) & n \\
\midrule\noalign{}
\endhead
\bottomrule\noalign{}
\endlastfoot
0 & 0.10 $\pm$ 0.00 & 4 \\
1,000 & 0.14 $\pm$ 0.04 & 4 \\
2,000 & 0.94 $\pm$ 0.17 & 4 \\
3,000 & 3.47 $\pm$ 1.04 & 4 \\
5,000 & \textbf{5.11 $\pm$ 0.95} (peak) & 4 \\
6,000 & 4.31 $\pm$ 1.57 & 4 \\
7,000 & 4.65 $\pm$ 0.84 & 4 \\
10,000 & 4.60 $\pm$ 0.48 & 4 \\
20,000 & 3.07 $\pm$ 0.91 & 4 \\
30,000 & 2.46 $\pm$ 0.28 & 4 \\
36,000 & 2.49 $\pm$ 0.12 & 4 \\
\end{longtable}
}

\emph{Table 3: 4-seed compositional fragility evolution. The std collapses
from 1.57 (step 6K) to 0.12 (step 36K); at the late plateau the
four seeds converge tightly.}

The compositional probe reproduces the qualitative pattern (accuracy
plateaus by step \textasciitilde5K; mean critical noise continues evolving through
step 36K) and shows its own quantitatively distinct long-term shape:
fragility rises through step 5K alongside accuracy onset (4-seed
mean 0.10 $\to$ 5.11), then declines through step 30K (5.11 $\to$ 2.46) and
holds. To verify the post-step-7K decline is replicable rather than
a single-seed artifact, we apply a pre-registered decision rule: the
decline counts as real if 4-seed mean critical noise drops by $\ge$ 1.0
between step 7K and step 30K \emph{and} seed-to-seed std at both endpoints
is smaller than the gap. Realized values: gap = 4.65 $-$ 2.46 = 2.19
($\ge$ 1.0 \checkmark{}), max endpoint std = 0.84 (\textless{} 2.19 \checkmark{}). Both pass with
substantial margin; the post-step-7K decline is a stable property
across the four split seeds.

Two non-exclusive readings of the diverging long-term direction
(7B / 32B replication disambiguates both): a \emph{mechanism reading} (as training
continues on text that does not specifically reinforce compositional
moral integration, the compositional representation drifts toward
brittleness while standard-probe representations are continually
reinforced by moralized vocabulary density) and a \emph{probe-ceiling
reading} (fragility at the 0.77 operating point has less headroom
than at 0.96, partly artifacting the difference). We state both in
\Cref{limitations} without commitment.

Numbers sources: \path|outputs/phase_c1/RESULTS.md| (1B standard probe),
\path|outputs/phase_b/| (7B corroboration),
\path|outputs/phase_c4_compositional/3seed/{aggregate_per_checkpoint,decision}.json|
(4-seed mean $\pm$ std and decision rule application),
\path|outputs/phase_c4_compositional/3seed/4seed_fragility_evolution.png|
(headline 4-seed plot).

\subsection{Data curation reshapes probe robustness, not probe accuracy}\label{data-curation-reshapes-probe-robustness-not-probe-accuracy}

LoRA \citep{hu2022lora} fine-tuning on three matched
corpora from the OLMo-2 1B step-1000 checkpoint (mid-transition, \textasciitilde80 \% peak probing
accuracy). Corpora: a 247K-token narrative-moral corpus (Aesop /
Grimm / Andersen), a 500K-token declarative-moral corpus
(template-expanded \path|MORAL_SEEDS|: ``Stealing is wrong''), and a 420K-token
general non-moral control (Darwin). Identical LoRA hyperparameters
(rank 16, alpha 32, q\_proj + v\_proj, lr 2e-4, batch 2, seq 1024,
1000 steps); standard moral probe + fragility every 100 LoRA steps.

\textbf{Probing accuracy is identical across conditions.} Final peak
accuracy at LoRA step 1000: narrative 0.740, declarative 0.750,
general control 0.750, all within 1 pp across very different training
data. The accuracy metric returns no signal for which corpus produces
what kind of representational change.

\textbf{Fragility profiles are condition-specific (the main result).}
Final mean critical noise: narrative 7.38, declarative 5.63, general
control 6.94. The per-layer breakdown separates the conditions:
narrative and general control show fragility dips at 6 and 7 of 16
layers respectively; the declarative condition shows dips at \textbf{10 of
16 layers}, creating a broadly more fragile representation than
either natural-text condition. \begin{figure}[t]
  \centering
  \includegraphics[width=\linewidth]{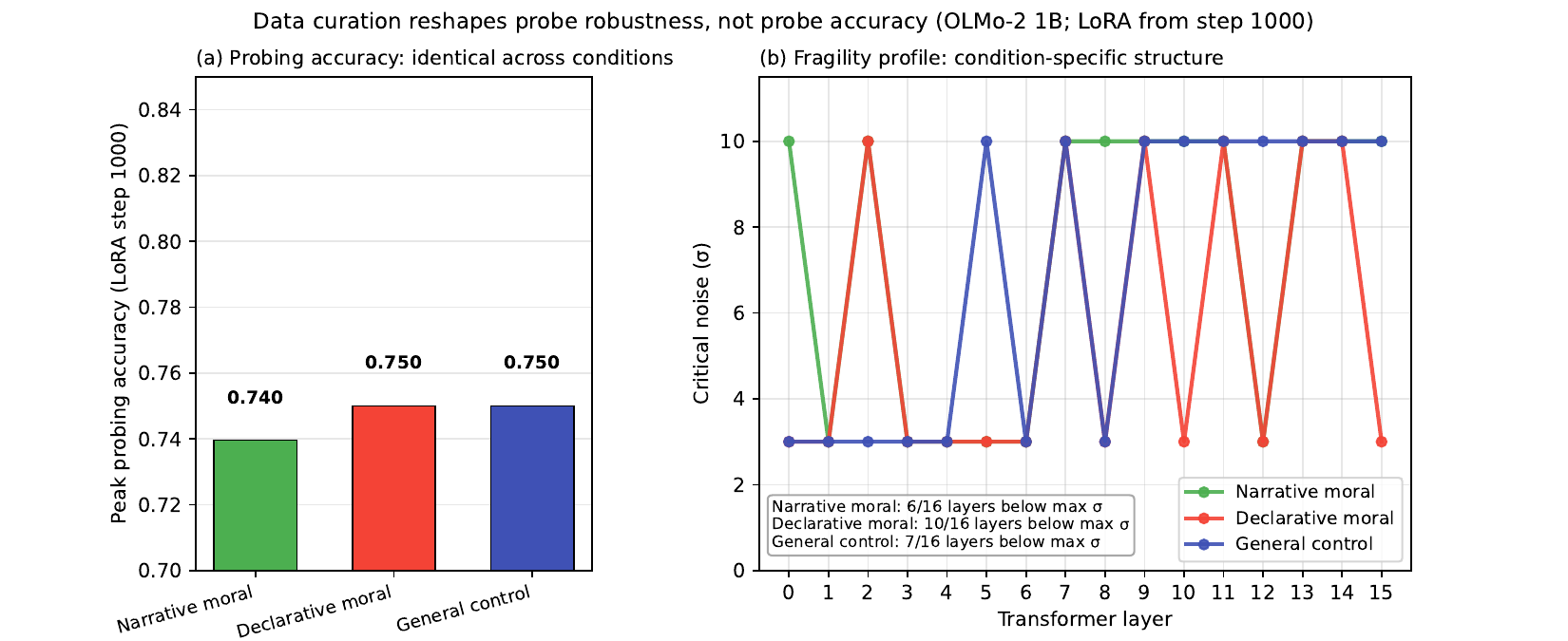}
  \caption{Data curation reshapes probe robustness, not probe accuracy (OLMo-2~1B; LoRA fine-tuning from step~1000 on three matched corpora). (a)~Final probing accuracy is near-identical across narrative-moral, declarative-moral, and general-text control conditions (0.740~/~0.750~/~0.750). (b)~Per-layer critical noise is condition-specific: declarative-moral training produces diffuse fragility across 10 of 16 layers (mean~$\sigma^*$=5.63 vs.~7.38/6.94 for natural-text conditions).}
  \label{fig:data-curation}
\end{figure}

\textbf{Figure~\ref{fig:data-curation}} plots all three per-layer profiles plus the three identical accuracy bars: same accuracy,
different fragility.

\textbf{Training loss is decoupled from representational change.}
Declarative loss drops 5.5 $\to$ 1.0 (template memorization), narrative
4.3 $\to$ 4.0, control 5.0 $\to$ 4.4. The condition with the deepest loss
reduction is the same condition with the most diffuse fragility; the
two with the shallowest loss reductions retain more robust
representations. The model is learning declarative templates as
surface text patterns without that learning translating into either
accuracy gains or robust representational structure.

\textbf{Why diffuse fragility under declarative training?} The declarative
corpus consists of repeated syntactic templates (``X is wrong'', ``Y
is immoral'') that the model memorizes easily (loss $\to$ 1.0). This
memorization creates narrow pattern-matching features across
multiple layers, features whose probe accuracy is maintained by
low-margin decision boundaries that collapse under small noise.
Narrative and general-control conditions produce fewer fragile
layers because natural text has no repeated syntactic template;
moral content in Aesop's fables is embedded in diverse narrative
structures, forcing the probe to rely on distributed features that
tolerate noise. The declarative fragility pattern is consistent
with the \Cref{probing-accuracy-saturates-fragility-doesnt} finding that early layers grow progressively more
brittle over training; declarative LoRA produces a diffuse
version of a vulnerability pattern that pre-training produces as
a layer-depth gradient.

This is direct evidence for the methodological thesis in a
controlled setting: same data, same probe; accuracy returns no
signal, fragility separates the conditions.

\textbf{Why the fragility is diffuse rather than localized.} The diffuse
pattern, fragility across 10 of 16 layers (mean $\sigma$* = 5.63 vs.
7.38 / 6.94) rather than a single dramatic dip, has a straightforward
mechanistic explanation. When the probing dataset controls for
animacy and register confounds (\Cref{standard-minimal-pair-datasets}), the probe detects actual
moral features at every layer rather than exploiting shortcuts like
``is this about a person or a circuit?'' that survive Gaussian
perturbation easily. When declarative templates are memorized, the
model's moral representations throughout the network become
template-dependent: narrow-margin features that collapse under
noise at multiple layers rather than just one. Template memorization
does not corrupt one layer; it degrades the network's moral
representations broadly. This sensitivity to dataset quality
confirms the importance of the validation methodology described
in \Cref{standard-minimal-pair-datasets}: probing datasets that contain animacy or register
shortcuts will systematically underestimate fragility at layers
where the probe exploits those shortcuts rather than moral content.

\textbf{Scale sensitivity of the \Cref{data-curation-reshapes-probe-robustness-not-probe-accuracy} ordering (v2 addendum).} Because \Cref{data-curation-reshapes-probe-robustness-not-probe-accuracy}
compares three separately fine-tuned models, the raw critical-noise
confound of \S4.4 (lower activation scale reads as higher fragility at
fixed absolute noise) can act across conditions, not only across layers.
The three conditions do not sit at identical activation scale: the
declarative model's per-layer activation RMS runs about 10\% below the two
natural-text conditions on average and 15--20\% below at deep layers (mean
RMS 2.21 vs 2.42 and 2.45). We therefore re-ran the exact cell (step-1000
checkpoint, identical corpora and LoRA settings, declarative memorized to
loss 1.01) under a scale-controlled estimator: an analytic critical noise
$\sigma$* computed in closed form from the frozen linear probe's margin
distribution (no grid, no cap), read both in raw activation units and
after per-layer RMS normalization (unit-RMS, an SNR-matched noise scale).
The coarse five-point grid used for the values above censors most layers,
and a naive RMS-normalization on that grid manufactures a spurious sign
reversal that the analytic estimator does not; that reversal is a
censoring artifact, not a scale effect.

The ordering survives scale-matching. Declarative remains the most fragile
condition under both readings (analytic raw $\sigma$* 2.90 vs 4.11 and 4.63;
SNR $\sigma$* 1.37 vs 1.75 and 1.87), and the paired bootstrap difference
$\sigma$*(narrative) $-$ $\sigma$*(declarative) excludes zero either way (raw $\Delta$ = 1.34,
95\% CI {[}0.24, 2.47{]}; SNR $\Delta$ = 0.44, 95\% CI {[}0.01, 0.91{]}). The declarative
condition is genuinely more fragile, not merely lower-scale. The scale
correction is not negligible: RMS normalization removes about two-thirds
of the raw analytic gap, so the scale-matched effect is roughly three
times smaller than the raw difference and only marginally clears zero. The
ordering is robust; the magnitude is scale-sensitive. The grid $\sigma$* values
reported above (5.63 / 6.94 / 7.38) are the coarse-grid estimate of the
same ordering and overstate the condition separation on two counts, grid
snapping and activation scale; the RMS-normalized SNR $\sigma$* with its paired
difference-CI is the quantity to cite. This is the cross-condition analog
of the \S4.4 cross-layer correction.

Numbers source: \path|outputs/phase_c_tier2/c3/RESULTS.md|,
\path|outputs/phase_c_tier2/c3/{narrative,declarative,general}_moral.json|
(per-layer fragility), and \path|outputs/phase_c_tier2/c3/confound_analytic/|
(the scale-controlled analytic re-analysis and paired difference-CIs).

\subsection{Scale-normalized fragility: a confound in raw critical noise}\label{scale-normalized-fragility-a-confound-in-raw-critical-noise}

Raw critical noise injects Gaussian noise in raw activation units. That
makes it sensitive to a confound we now isolate and correct: per-layer
activation scale. On OLMo-2 1B the root-mean-square of a layer's hidden
activations grows monotonically with depth, from \({\approx}0.13\) at the
early layers to \({\approx}1.0\) at the late layers (up to 1.66 at the
last layer), an \({\sim}8\times\) growth. A fixed raw \(\sigma\) is
therefore a much larger \emph{relative} perturbation at an early layer than
at a late one, so early layers will look fragile and late layers robust
\emph{regardless of representational quality}. The layer-depth gradient of
\Cref{probing-accuracy-saturates-fragility-doesnt} is exactly the shape this confound predicts.

\textbf{The control.} We re-evaluate fragility with the noise scaled to each
layer's activation scale: per layer and per checkpoint, standardize the
activations to unit RMS, retrain the probe, and inject noise on that
normalized scale (equivalently, scale \(\sigma\) by the per-layer RMS).
This measures probe margin in scale-free units. Two things change.

\emph{The layer-depth gradient mostly disappears.} The raw late/early
\(\sigma^*\) ratio grows over training to as much as \(14.7\times\) (step
10K) and is \(7.1\times\) at step 36K; under RMS normalization it stays
near \(2\times\) throughout (1.8--3.1\(\times\)) and the late \(\ge\) mid
\(\ge\) early ordering fails outright at 8 of 37 checkpoints. The
gradient is largely activation scale, not robustness. We do not claim
the residual \({\sim}2\times\) as a genuine robustness gradient. RMS
normalization equalizes per-layer scale but not covariance shape, so a
fixed isotropic perturbation still interacts differently with layers
whose activation covariances differ in anisotropy and effective
dimensionality, which can leave a residual ratio near \(2\times\) that is
not encoding robustness. Establishing a genuine residual gradient would
require a whitened or participation-ratio-matched control (noise
injected in each layer's own whitened basis), which we do not run here.

\emph{The post-saturation evolution mostly disappears.} Raw mean \(\sigma^*\)
declines from 18.3 (step 4K) to 4.7 (step 36K) after accuracy
saturates; RMS-normalized mean \(\sigma^*\) is flat over the same span
(13.8 $\to$ 15.0, slope \(+0.05\) per 1K steps). The compositional probe
behaves the same way: across all 37 checkpoints its raw mean \(\sigma^*\)
declines 6.6 $\to$ 1.8 after saturation while the RMS version is flat
(4.1 $\to$ 5.9).

\begin{figure}[t]
\centering
\includegraphics[width=\linewidth]{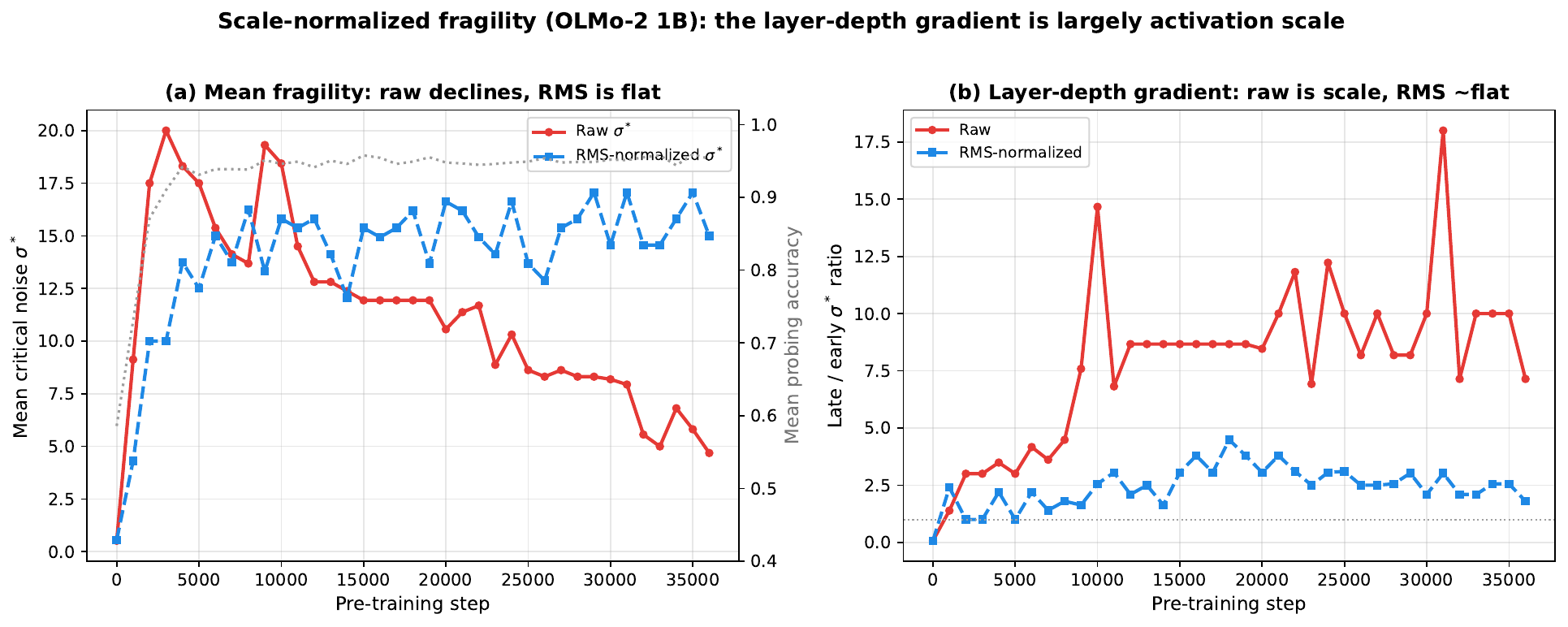}
\caption{Scale-normalized fragility on OLMo-2 1B. (a) Mean critical noise over training: raw $\sigma^*$ (red) rises then declines after accuracy (gray) saturates, while RMS-normalized $\sigma^*$ (blue) is flat. (b) The late/early $\sigma^*$ ratio: the raw layer-depth gradient grows to 7--15$\times$, but under RMS normalization it stays near 2$\times$. The cross-layer gradient and the post-saturation evolution are largely activation scale.}
\label{fig:rms_control}
\end{figure}

\textbf{What survives.} The confound is strictly \emph{cross-layer}: it arises
because different layers have different activation scales. It does not
touch \emph{within-layer} comparisons, where the scale is held fixed by
construction. The \Cref{data-curation-reshapes-probe-robustness-not-probe-accuracy} data-curation result is exactly such a
comparison: narrative, declarative, and general-control conditions are
contrasted at each matched layer, on the same base checkpoint with the
same corpus size and LoRA hyperparameters, so their per-layer scales
are near-identical and the declarative-vs-natural fragility separation
is not a scale artifact. The same holds for cross-checkpoint
comparisons at a fixed layer and cross-architecture comparisons at
matched layers. Raw \(\sigma^*\) also retains a direct reading on its own
terms, even cross-layer: it is \emph{practical perturbation sensitivity},
the absolute noise a representation tolerates before the probe
degrades, which is the relevant quantity for adversarial-robustness or
fine-tuning-stability questions. What it does not measure, cross-layer,
is encoding robustness independent of scale.

We therefore recommend \textbf{RMS-normalized \(\sigma^*\) for cross-layer
claims and raw \(\sigma^*\) for within-layer comparisons} (where scale
is controlled by design). This distinction is itself a contribution:
noise-injection probing is widely used, and the activation-scale
confound applies to any such method that compares across layers in raw
units. We develop this caution, with other interpretability-instrument
failure modes, in a companion methods note (in preparation).

Numbers source: \path|outputs/phase_c1_refragility/trajectory.json|
(seed-averaged extended-grid raw and RMS-normalized fragility for all
37 checkpoints) and \path|outputs/phase_c4_comp_refragility/trajectory.json|
(compositional probe).

\section{Discussion}\label{discussion}

\subsection{Semantic vs.~structural learning dynamics}\label{semantic-vs.-structural-learning-dynamics}

The \Cref{emergence-ordering} four-curve overlay (Figure 1) shows two distinct learning regimes
within a single training run on a single model. The standard moral
and sentiment probes, both single-token-swap minimal-pair tasks,
emerge as sharp, step-like sigmoidal transitions: each crosses from
chance to its plateau within a single 1K-step interval at onset, then
flattens. (At 1K sampling we resolve transitions that are step-like
at this resolution; we do not claim true discontinuity.) The compositional moral and syntax probes, both tasks that
require multi-token integration to determine the label, rise more
gradually, with no equally sharp inflection point.

The cleanest hypothesis to organize this dichotomy: \textbf{phase-transition
dynamics emerge when a feature can be acquired through local lexical
or distributional statistics (the model ``discovers'' the feature in
a discrete jump as soon as it has enough samples to distinguish the
relevant lexemes), while gradual emergence indicates features that
require integrating positional, attentional, or compositional
relationships across multiple tokens, which the model cannot acquire
in a single step from local lexical statistics alone.} Under this
reading the standard moral probe (single-lexeme swap), sentiment
(single-adjective swap), and similar lexically-localized tasks all
share the phase-transition mechanism; the compositional probe
(multi-token integrated swap) and syntax (positional well-formedness)
share the gradual-emergence mechanism. The 0.20 plateau gap (\Cref{emergence-ordering})
between the two regimes (single-token-statistics tasks saturating
near 0.97, multi-token-integration tasks near 0.77) is consistent
with this reading: features that can be cleanly read off single-token
distributional statistics in mean-pooled hidden states should reach
higher linear separability than features that require recovering
multi-token interactions from a pooling operation that discards
positional information.

The dichotomy connects to \citeauthor{power2022grokking}'s \citeyearpar{power2022grokking} grokking literature
(sudden phase transitions on algorithmic tasks), which has largely
focused on the \emph{cause} of phase transitions; our results suggest the
\emph{taxonomy} of which capabilities should and should not exhibit them.
The formal information-theoretic argument is its own paper.

\subsection{Why fragility succeeds where accuracy saturates}\label{why-fragility-succeeds-where-accuracy-saturates}

Probing accuracy is a thresholded, capped, top-end metric: once
linear separability is good enough, accuracy hits ceiling and stops
returning information about underlying representational change.
Fragility is structurally different: it is sensitive to both the
\emph{margin} of separability (outputs near the decision boundary flip
under small noise) and the \emph{redundancy} of representation (features
encoded in many hidden-space directions tolerate noise that collapses
any one). It does not separately identify their contributions, and a
low critical noise can also reflect activation-scale changes,
representational anisotropy, or probe-training instability rather than
margin or redundancy alone. Both margin and redundancy continue to
evolve after accuracy saturates because both are functionals of
representation \emph{geometry} rather than end-to-end classification
accuracy. Two cautions follow from \S4.4, however. First, much of the
raw cross-layer and cross-checkpoint movement of critical noise is
activation scale, not margin or redundancy: it must be read in
RMS-normalized units for cross-layer claims. Second, the place where
fragility demonstrably separates representations that accuracy cannot
is the \emph{scale-controlled} comparison: the \Cref{data-curation-reshapes-probe-robustness-not-probe-accuracy} data-curation contrast,
where three corpora with identical probing accuracy produce distinct
fragility fingerprints at matched layers, on a fixed checkpoint, with
activation scale held constant by construction. That is the clean
demonstration of the methodological thesis. The broader contribution is
therefore twofold: critical noise is a probe-side metric that keeps
moving after accuracy saturates, and, just as important, we delimit
\emph{when} it is measuring representation rather than scale, raw \(\sigma^*\)
within-layer and RMS-normalized \(\sigma^*\) across layers. Fragility is
not a moral-domain-specific tool but a methodological one for any binary
probing task that hits the accuracy ceiling, provided the scale
confound is controlled.

\subsection{Limitations}\label{limitations}

\textbf{Staged emergence bounds the standard probe.} The
standard moral probe measures something closer to ``moralized
vocabulary becomes linearly separable from neutral vocabulary'' than
``moral reasoning emerges.'' The compositional probe (\Cref{emergence-ordering})
established this emerges in stages: morally loaded words
at step 1K, compositional moral integration at
step 5K, syntactic competence at step 6K, not a binary
in-or-out distinction. Both onsets are real findings. Neither of
them is ``moral reasoning at step 1K''; both are bounded claims about
what a linear probe can recover from mean-pooled hidden states at
each step.

\textbf{Compositional probe partial scope.} The compositional probe
addresses \emph{whether the moral signal lives in single-token vs.
multi-token features}; it does not address \emph{whether the model
represents moral concepts in any deeper functional sense}:
counterfactual sensitivity to moral reframing, generalization to
novel moral structures not in pre-training data, behavioral
consistency under adversarial probing. The compositional probe is a
strictly stronger lexical-accessibility ablation than the standard
probe; it is not a moral-reasoning probe. Stronger probes for
deeper moral capacities are out of scope for this paper.

\textbf{Two related questions disambiguate at scale (7B / 32B
replication).} First, the \Cref{emergence-ordering} plateau coincidence (compositional
$\approx$ syntax $\approx$ 0.77 vs.~standard moral / sentiment $\approx$ 0.97) may reflect
a 1B-model ceiling on compositional / structural encoding or a
probe-side ceiling under mean-pooled linear probing. Second, the
\Cref{probing-accuracy-saturates-fragility-doesnt} 4-seed compositional fragility decline through steps 7K-30K
(4.65 $\to$ 2.46), opposite to the standard probe's late-training
hold, admits both a \emph{mechanism reading} (compositional
representations drift toward brittleness as training continues on
text that does not specifically reinforce them) and a \emph{probe-ceiling
reading} (fragility at the 0.77 operating point has less headroom
than at 0.96, partly artifacting the difference). Both readings
predict different scaling behavior: under the mechanism reading the
decline tracks training-text distribution rather than scale, under
the probe-ceiling reading it attenuates as scale lifts the
operating point. Repeating \Cref{emergence-ordering} and \Cref{probing-accuracy-saturates-fragility-doesnt} at 7B and 32B disambiguates
both. Either outcome refines the staged-emergence finding without overturning
it.

\textbf{Single model family.} All findings are on OLMo-2 1B and OLMo-3
7B. Generalization to other architectures and training recipes is
open.

\textbf{Single language.} All probing datasets are English; pretraining
data for both target models is dominantly English. Cross-lingual
generalization of both the staged-emergence finding and the
fragility-resolves-what-accuracy-misses pattern is open.

\textbf{Raw-$\sigma$ fragility and the scale confound (resolved in \S4.4).} We add
Gaussian noise in raw activation units (\Cref{fragility-testing}), which conflates probe
margin with activation scale: hidden-state RMS grows \textasciitilde8$\times$ from early to
late layers, so a fixed raw $\sigma$ is a larger relative perturbation early
than late. \S4.4 runs the RMS-normalized control and finds that the
\Cref{probing-accuracy-saturates-fragility-doesnt} layer-depth gradient and its cross-checkpoint evolution are largely
this scale effect (the late/early $\sigma$* ratio falls from \textasciitilde7--15$\times$ to \textasciitilde2$\times$,
and the post-saturation decline flattens). The confound is strictly
cross-layer, so it leaves within-layer comparisons intact, including the
\Cref{data-curation-reshapes-probe-robustness-not-probe-accuracy} declarative-vs-natural separation (contrasted at matched layers on
a shared checkpoint, hence at fixed scale). We therefore recommend
RMS-normalized $\sigma$* for cross-layer claims and raw $\sigma$* for within-layer
comparisons, and we read raw cross-layer $\sigma$* only as practical
perturbation sensitivity, not as scale-independent encoding robustness.
The one cell we did not re-run under RMS is the LoRA experiment itself;
its within-layer construction guarantees scale is controlled, but an
explicit RMS replication (saving adapters for reuse) is left to the
multi-seed LoRA extension.

\textbf{Noise grid.} Critical noise is read off the extended $\sigma$ grid
\(\{0.1, 0.3, 1, 3, 10, 30, 100\}\) (\Cref{fragility-testing}), which lifts the late-layer
band off the old $\sigma$=10 cap (late layers reach 30--44 mid-trajectory in
Table 2). Layers that never drop below $\tau$ even at $\sigma$=100 are censored at
100; this is rare on the 1B trajectory.

\textbf{Foundation-specific scope.} The standard moral dataset's six MFT
foundations show staggered emergence; all six stabilize by step 3K
(Appendix A). The emergence ordering is sensitive to dataset
construction choices, which confirms the importance of dataset quality
methodology in probing studies; findings that depend on specific
pairs rather than the property of interest are artifacts, not
discoveries. The compositional dataset's 200 pairs are categorized by
construction pattern (motive / target / consequence / role) rather
than by MFT foundation; a foundation-stratified compositional probe
(parallel to the foundation-specific standard probe in Appendix A)
would tell us whether different foundations acquire compositional
encoding at different steps. Out of scope for this paper but a
natural extension.

\section{Conclusion}\label{conclusion}

Probing accuracy is the wrong instrument for studying training
dynamics. Whatever continued representational change a language model
undergoes during pre-training after a property becomes linearly
decodable (and \Cref{results} shows there is a great deal of such change) does
not register on a metric that has already saturated. The methodological
contribution of this paper is to take this saturation as a fixed
feature of probing accuracy and add a complementary metric (per-layer
critical noise, which we call \emph{fragility}) that recovers the missing
resolution. The contribution is not specific to moral representations
or to pre-training trajectories; we have established the
fragility-resolves-what-accuracy-misses pattern on the standard moral
probe (\Cref{probing-accuracy-saturates-fragility-doesnt}) and on the compositional moral probe across four random-
seed splits (\Cref{probing-accuracy-saturates-fragility-doesnt}, replication). We expect the methodology to extend
to any binary probing-based
investigation of neural network representations where the operating
point is high enough to saturate the standard accuracy curve.

The science findings earn the methodology its keep. The most
consequential of them is the \emph{staged emergence} the paper
establishes by adding the compositional moral probe alongside the
standard moral, sentiment, and syntax probes: morally loaded
words are decoded at step 1K, compositional moral
integration at step 5K, syntactic competence at step 6K. The
strongest one-sentence summary of the moral probe's step-1K onset,
that ``moral encoding emerges within the first 5\% of pre-training,''
overstates what the standard probe recovers; the staged framing
(``morally loaded words at 1K, compositional moral
integration at 5K, syntactic competence at 6K'') is the honest
reading. We treat this as an existence proof that probing-claims
about pre-training emergence should default to lexical-accessibility
hedges until a compositional-or-stronger ablation is run.

Two open questions remain. First, does the
compositional plateau at $\approx$0.77 lift with model scale (7B / 32B
replication of \Cref{emergence-ordering})? The plateau coincidence is a probe-side
property at 1B; scale disambiguates whether the ceiling is the model
or the instrument. Second, does the fragility-resolves-what-accuracy-
misses pattern extend beyond the probing investigations we have run
(foundation-stratified compositional probing, counterfactual-
sensitivity probing, linguistic-property probes, factual-recall
probes, persona-feature probes)? If fragility resolves dynamics
that accuracy misses across this broader set, the methodology is a
general-purpose tool for alignment-during-pre-training research; if
not, the conditions under which it generalizes are themselves the
contribution of follow-up work.

The pre-training window does more work than the standard
interpretability instrument shows. Adding a fragility readout
recovers a substantial fraction of that work. The staged emergence
is what we recover when we apply the methodology to moral
representations specifically; the broader claim is that the same
methodology will recover analogous structure wherever probing-based
investigations of pre-training currently hit the accuracy ceiling.

\begin{ack}
This work made extensive use of Anthropic's Claude (the Claude Code agent on
Opus~4.6, 4.7, 4.8 and Fable~5) for code scaffolding, experimental scripts, and
prose drafting. The author retains responsibility for experimental design, all
scientific claims, and final wording.
\end{ack}

\bibliography{references}

\appendix
\newpage
\begin{center}
  \rule{0.5\linewidth}{0.4pt}\\[0.6em]
  {\Large\bfseries Appendices}\\[0.25em]
  {\small Supplementary material. Sections referenced from the main
   paper as ``Appendix A''--``Appendix E''.}\\[0.4em]
  \rule{0.5\linewidth}{0.4pt}
\end{center}
\vspace{0.5em}

\section{Foundation emergence}\label{appendix-a.-foundation-emergence}

The standard moral probe's 240 minimal-pair dataset is balanced
across the six Moral Foundations Theory categories \citep{haidt2012righteous,graham2013mft} at 40 pairs per foundation: care/harm,
fairness/cheating, loyalty/betrayal, authority/subversion,
sanctity/degradation, and liberty/oppression. The foundation-
stratified probe (\texttt{FoundationSpecificProbe}) trains a separate linear
classifier per foundation, allowing per-foundation onset and plateau
analysis across the same 37 OLMo-2 1B early-training checkpoints
used in \Cref{results}.

\textbf{Headline finding.} All six foundations stabilize (reach 100\% peak
probing accuracy) by step 3K, within the same window where the
aggregated standard moral probe onsets. Per-foundation onset
differences are within sampling noise: the per-foundation peak
accuracies are sixteenths on a 16-example test, single seed (e.g.
step-1K authority 100\% = 16/16 vs.~care 87.5\% = 14/16, a two-example
gap), so we do not read the apparent step-1K / 2K / 3K ordering as a
reliable per-foundation sequence.

{\def\LTcaptype{none} 
\begin{longtable}[]{@{}
  >{\raggedright\arraybackslash}p{(\linewidth - 12\tabcolsep) * \real{0.24}}
  >{\raggedleft\arraybackslash}p{(\linewidth - 12\tabcolsep) * \real{0.09}}
  >{\raggedleft\arraybackslash}p{(\linewidth - 12\tabcolsep) * \real{0.09}}
  >{\raggedleft\arraybackslash}p{(\linewidth - 12\tabcolsep) * \real{0.09}}
  >{\raggedleft\arraybackslash}p{(\linewidth - 12\tabcolsep) * \real{0.09}}
  >{\raggedleft\arraybackslash}p{(\linewidth - 12\tabcolsep) * \real{0.09}}
  >{\raggedleft\arraybackslash}p{(\linewidth - 12\tabcolsep) * \real{0.31}}@{}}
\toprule\noalign{}
\begin{minipage}[b]{\linewidth}\raggedright
Foundation
\end{minipage} & \begin{minipage}[b]{\linewidth}\raggedleft
Step 0
\end{minipage} & \begin{minipage}[b]{\linewidth}\raggedleft
Step 1K
\end{minipage} & \begin{minipage}[b]{\linewidth}\raggedleft
Step 2K
\end{minipage} & \begin{minipage}[b]{\linewidth}\raggedleft
Step 3K
\end{minipage} & \begin{minipage}[b]{\linewidth}\raggedleft
Step 6K
\end{minipage} & \begin{minipage}[b]{\linewidth}\raggedleft
First step at 100\%
\end{minipage} \\
\midrule\noalign{}
\endhead
\bottomrule\noalign{}
\endlastfoot
authority/subversion & 68.8\% & 100\% & 100\% & 100\% & 100\% & 1K \\
care/harm & 68.8\% & 87.5\% & 100\% & 93.8\% & 100\% & 2K \\
fairness/cheating & 75.0\% & 75.0\% & 100\% & 100\% & 100\% & 2K \\
sanctity/degradation & 68.8\% & 75.0\% & 93.8\% & 100\% & 100\% & 3K \\
loyalty/betrayal & 62.5\% & 62.5\% & 93.8\% & 100\% & 100\% & 3K \\
liberty/oppression & 68.8\% & 93.8\% & 87.5\% & 100\% & 100\% & 3K \\
\end{longtable}
}

\emph{Per-foundation peak probing accuracy at key OLMo-2 1B early-training
checkpoints. Numbers source: \path|outputs/phase_c1/| foundation-specific
probe results.}

\textbf{All six foundations stabilize by step 3K.} During development,
an earlier dataset with weaker neutral-pair quality produced
unstable liberty/oppression encoding; the instability resolved when
neutral sentences that inadvertently carried moral content were
removed. This sensitivity to dataset quality confirms the
importance of the validation methodology described in \Cref{standard-minimal-pair-datasets}.
The apparent per-foundation ordering (authority at step 1K, care and
fairness at step 2K, the remaining three by step 3K) is within the
sampling noise of a 16-example, single-seed test; the reliable
statement is that all six stabilize by step 3K.

Numbers source: \path|outputs/phase_c1/| (1B trajectory) and
\path|outputs/phase_b/b3_foundation_emergence.png| (7B comparison).

\section{Causal-probing divergence}\label{appendix-b.-causal-probing-divergence}

The \Cref{methodology} methodology distinguishes \emph{probing accuracy} (how linearly
decodable a property is from a layer's hidden states) from \emph{causal
contribution} (how strongly intervening on a layer's hidden states
changes the model's downstream behavior on the property). The two
are conceptually separate (the layer where information is \emph{stored}
may be different from the layer where information is \emph{used}), but
the distinction is rarely operationalized in moral-representation
work, which typically reports probing accuracy and stops there.

We applied the \texttt{MoralCausalTracer} benchmark (an adaptation of Meng
et al.'s 2022 ROME causal-tracing methodology to the moral domain)
on three OLMo-3 7B checkpoints (early, mid, final) using the same
240-pair standard moral dataset as the probing analysis. OLMo-3 7B has
32 transformer layers, so layer indices run 0--31. The headline finding:

{\def\LTcaptype{none} 
\begin{longtable}[]{@{}
  >{\raggedleft\arraybackslash}p{(\linewidth - 6\tabcolsep) * \real{0.1644}}
  >{\raggedleft\arraybackslash}p{(\linewidth - 6\tabcolsep) * \real{0.2603}}
  >{\raggedleft\arraybackslash}p{(\linewidth - 6\tabcolsep) * \real{0.2740}}
  >{\raggedleft\arraybackslash}p{(\linewidth - 6\tabcolsep) * \real{0.3014}}@{}}
\toprule\noalign{}
\begin{minipage}[b]{\linewidth}\raggedleft
Checkpoint
\end{minipage} & \begin{minipage}[b]{\linewidth}\raggedleft
Peak causal layer
\end{minipage} & \begin{minipage}[b]{\linewidth}\raggedleft
Peak probing layer
\end{minipage} & \begin{minipage}[b]{\linewidth}\raggedleft
Mean indirect effect
\end{minipage} \\
\midrule\noalign{}
\endhead
\bottomrule\noalign{}
\endlastfoot
Step 0 & 5 & 0 & 0.01 \\
Step 705K & 5 & 19 & 7.84 \\
Step 1,414K & 0 & 10 & 7.95 \\
\end{longtable}
}

\emph{Numbers source (exact files under \path|outputs/phase_b/|): peak causal
layer and the layer-mean indirect effect from
\path|step_{0000000,0705000,1413814}/moral_causal_tracer_allenai_OLMo-3-1025-7B.json|
(\path|peak_causal_layer|, and the mean over \path|mean_indirect_effect_by_layer|);
peak probing layer from
\path|step_{0000000,1413814}/layer_wise_moral_probe_allenai_OLMo-3-1025-7B.json|.
The causal subset (steps 0 / 705K / 1,414K) did not include a
co-located probe at step 705K, so that row's probing layer (19) is
taken from the nearest probed checkpoint, step 668K. By mid-training
many layers saturate at \textasciitilde1.0 probing accuracy, so the single
argmax ``peak probing layer'' is unstable across probe seeds; we read it
as ``mid-network'' rather than as an exact layer.}

Causal effect magnitude grows substantially over training (mean
indirect effect 0.01 $\to$ 7.95), and the peak causal layer migrates
from layer 5 $\to$ layer 5 $\to$ layer 0. The peak probing layer sits in
mid-network throughout (layer 0 at init, then layers 19 / 10 once
accuracy saturates). At the final checkpoint, the gap between the
mid-network layer that most-strongly \emph{encodes} moral information
(layer 10) and the early layer that most-strongly \emph{influences}
downstream moral-relevant generation (layer 0) is 10 layers; the two
metrics identify opposite ends of the network.

This is consistent with a ``storage vs.~use'' picture of moral
representation in transformer language models: moral information is
\emph{stored} in mid-network layers (where probing recovers it cleanly)
and \emph{used} in early layers (where intervening on it most-strongly
moves the model's downstream output). The two facts are
representational properties of the same model that probing alone
cannot recover.

This appendix is an uncontrolled, exploratory illustration of the \Cref{why-fragility-succeeds-where-accuracy-saturates}
fragility-as-richer-functional point: probing accuracy is one
functional of the representation geometry; causal contribution is
another; fragility is a third. All three keep evolving through
training, often in different directions, and a complete picture of
how a representation evolves needs all three. We do not develop the
storage-vs-use finding as a body contribution because (a) it is not
specific to moral representation and (b) it has its own
methodological complications (causal-tracing sensitivity to
intervention magnitude, choice of decoder probe) that warrant their
own paper.

\section{Standard moral probe validity controls}\label{appendix-c.-standard-moral-probe-validity-controls}

Three controls are standard for linear-probing studies that claim to
measure something beyond surface vocabulary:

\begin{enumerate}
\def\labelenumi{\arabic{enumi}.}
\tightlist
\item
  \textbf{Leave-lexeme-out splits.} Train the probe with all pairs
  containing a target lexeme (e.g.~all ``betray'' pairs) held out;
  evaluate on those held-out pairs. Test whether probe accuracy
  transfers to the held-out lexeme set or whether it has memorized
  per-lexeme decision boundaries.
\item
  \textbf{Paraphrase transfer.} Generate paraphrases of test-set pairs
  that preserve moral content but vary surface form; evaluate the
  probe trained on the original test-set pairs on the paraphrased
  versions. Test whether the probe recovers the moral signal under
  surface variation or whether it is reading per-pair surface
  features.
\item
  \textbf{Adversarial lexical swap.} Construct adversarial pairs where
  a surface feature the probe might be using (sentence length,
  position of the moral lexeme, presence of specific function
  words) is decoupled from the moral label. Test whether the probe
  accuracy degrades on the adversarial set.
\end{enumerate}

The compositional moral probe (\Cref{compositional-moral-probing-dataset}) addresses the strongest version
of the ``your probe is just reading moralized vocabulary'' concern \emph{by
construction}: pairs share the morally-loaded action verb between
halves and differ only in tokens that carry limited moral signal in
isolation (unigram lexical floor 0.63). The probe's compositional
encoding is established directly by leave-construction-out transfer
(0.85 hidden-state vs.~0.60 bag-of-words) and by decoding 0.20-0.28
above the per-construction lexical floor (\Cref{compositional-moral-probing-dataset}, \Cref{emergence-ordering}).

We rest the validity argument on this compositional result rather than
on the three controls above. Leave-lexeme-out and adversarial-lexical-swap
both ask whether the probe reads moral content beyond a specific surface
cue; the compositional probe answers the stronger question directly, by
holding the morally-loaded verb fixed across both halves of every pair and
showing that moral valence still decodes, and transfers, from context
alone. A probe reading lexeme identity or any single surface feature could
not pass leave-construction-out transfer at 0.85. The compositional result
therefore subsumes what the leave-lexeme-out and adversarial-swap controls
would test, and paraphrase transfer is the one remaining surface-variation
check we leave to future work.

\section{Compositional probe pair list and per-category breakdown}\label{appendix-d.-compositional-probe-pair-list-and-per-category-breakdown}

The 200-pair compositional moral minimal-pair dataset constructed
for \Cref{compositional-moral-probing-dataset} and \Cref{emergence-ordering} is released in
\path|deepsteer/datasets/compositional_moral_pairs.py| as the
\path|COMPOSITIONAL_MORAL_PAIRS| constant. Each pair is a
\texttt{(moral\_text,\ immoral\_text)} tuple; the file is plain Python with
inline comments grouping pairs by category.

\subsection{Per-category structure}\label{d.1-per-category-structure}

{\def\LTcaptype{none} 
\begin{longtable}[]{@{}
  >{\raggedright\arraybackslash}p{(\linewidth - 6\tabcolsep) * \real{0.2708}}
  >{\raggedright\arraybackslash}p{(\linewidth - 6\tabcolsep) * \real{0.2083}}
  >{\raggedleft\arraybackslash}p{(\linewidth - 6\tabcolsep) * \real{0.1458}}
  >{\raggedright\arraybackslash}p{(\linewidth - 6\tabcolsep) * \real{0.3750}}@{}}
\toprule\noalign{}
\begin{minipage}[b]{\linewidth}\raggedright
Index range
\end{minipage} & \begin{minipage}[b]{\linewidth}\raggedright
Category
\end{minipage} & \begin{minipage}[b]{\linewidth}\raggedleft
Count
\end{minipage} & \begin{minipage}[b]{\linewidth}\raggedright
Contrast pattern
\end{minipage} \\
\midrule\noalign{}
\endhead
\bottomrule\noalign{}
\endlastfoot
0-49 & \path|action_motive| & 50 & same action verb, motive differs \\
50-99 & \path|action_target| & 50 & same action, target descriptor differs \\
100-149 & \path|action_consequence| & 50 & same action, consequence framing differs \\
150-199 & \path|role_reversal| & 50 & same components, role/target/context determines valence \\
\end{longtable}
}

\subsection{Per-category content-only TF-IDF baseline}\label{d.2-per-category-content-only-tf-idf-baseline}

Pair-disjoint five-fold \texttt{GroupKFold} (each minimal pair held together
so its shared skeleton cannot leak across the train / test split),
\texttt{TfidfVectorizer} and \path|LogisticRegression(max_iter=1000)|, scored
orientation-invariantly as \texttt{max(acc,\ 1\ -\ acc)}:

{\def\LTcaptype{none} 
\begin{longtable}[]{@{}lr@{}}
\toprule\noalign{}
Category & TF-IDF separability (pair-disjoint) \\
\midrule\noalign{}
\endhead
\bottomrule\noalign{}
\endlastfoot
action\_motive & 0.64 \\
action\_target & 0.59 \\
action\_consequence & 0.61 \\
role\_reversal & 0.57 \\
\textbf{Overall (\path|min_df=2|)} & \textbf{0.63} \\
\end{longtable}
}

These are the \emph{lexical floor}: the bag-of-words separability that the
hidden-state transfer-and-lift analysis (\Cref{compositional-moral-probing-dataset}, \Cref{emergence-ordering}) is measured
against, not a stand-alone compositionality certificate. A
single-contrast-token minimal pair is lexically separable by
construction, so a low floor is neither expected nor required; what
establishes compositional encoding is that hidden-state probes transfer
across construction categories (0.85 vs.~0.60 bag-of-words) and decode
well above these per-construction floors. Pair-disjoint folds and
orientation-invariant scoring both matter: random folds leak the
shared pair skeleton, and scoring a leaky anti-correlated classifier
by raw accuracy understates separability. Numbers source:
\path|deepsteer.datasets.compositional_moral_pairs.content_separability_baseline|.

\subsection{Construction-gate verification}\label{d.3-construction-gate-verification}

\path|validate_compositional_dataset()| enforces five gates simultaneously
and all 200 pairs pass all five:

\begin{enumerate}
\def\labelenumi{\arabic{enumi}.}
\tightlist
\item
  \textbf{Length difference} $\le$ 2 alphabetic tokens per pair (max
  observed: 2)
\item
  \textbf{Length band} 8-20 alphabetic tokens per half (range observed:
  8-19)
\item
  \textbf{Content-word overlap} $\ge$ 0.60 (stopwords removed; matches
  \path|deepsteer.datasets.validation.validate_pairs| metric)
\item
  \textbf{Strong-valence blocklist}: zero tokens from a 47-word
  blocklist (\texttt{murder}, \texttt{torture}, \texttt{stole}, \texttt{assault}, etc.) on
  either side
\item
  \textbf{No exact duplicates} of either side across the 200 pairs
\end{enumerate}

The construction iterated through approximately five rewriting
passes to satisfy gate 3 (content-word overlap $\ge$ 0.60) alongside
the multi-word compositional contrast requirement. The two
constraints are in genuine tension: compositional contrasts
naturally require swapping multiple content words, while
$\ge$0.60 content overlap forces single- or two-word swaps within
otherwise-identical structure, and the published dataset reflects
the achievable balance.

\subsection{Representative samples per category}\label{d.4-representative-samples-per-category}

Eight representative pairs (two per category):

\textbf{action\_motive:}
- \emph{moral:} ``She lied to her parents to \textbf{protect} her younger brother today''
- \emph{immoral:} ``She lied to her parents to \textbf{humiliate} her younger brother today''
- \emph{moral:} ``He invited the new colleague to lunch to make her feel \textbf{welcome}''
- \emph{immoral:} ``He invited the new colleague to lunch to make her feel \textbf{unwelcome}''

\textbf{action\_target:}
- \emph{moral:} ``He gave the last loaf of bread to the \textbf{hungry} stranger at the door''
- \emph{immoral:} ``He gave the last loaf of bread to the \textbf{wealthy} stranger at the door''
- \emph{moral:} ``She offered her warm coat to the \textbf{shivering child} waiting in the rain''
- \emph{immoral:} ``She offered her warm coat to the \textbf{bored heir} waiting in the rain''

\textbf{action\_consequence:}
- \emph{moral:} ``He kept the secret about the surprise to keep his sister \textbf{safe} today''
- \emph{immoral:} ``He kept the secret about the surprise to keep his sister \textbf{hurt} today''
- \emph{moral:} ``She held onto the savings to \textbf{provide for her children's} future education''
- \emph{immoral:} ``She held onto the savings to \textbf{deprive her children of any} future education''

\textbf{role\_reversal:}
- \emph{moral:} ``The judge accepted the gift to free the \textbf{innocent} prisoner from prison''
- \emph{immoral:} ``The judge accepted the gift to free the \textbf{guilty} prisoner from prison''
- \emph{moral:} ``The senator added the amendment to \textbf{protect} the whistleblowers from retaliation''
- \emph{immoral:} ``The senator added the amendment to \textbf{expose} the whistleblowers to legal retaliation''

\subsection{Full pair list}\label{d.5-full-pair-list}

The complete 200-pair list is released in
\path|deepsteer/datasets/compositional_moral_pairs.py|. The dataset is
deterministic and API-free; reproducing the results in \Cref{emergence-ordering} requires
only loading the constant and the \texttt{LayerWiseMoralProbe} /
\texttt{MoralFragilityTest} infrastructure released with the paper.

A camera-ready version of this appendix will inline the full pair
list (one pair per row in a table) for archival completeness if the
venue's appendix length allows; otherwise the published code
repository is the canonical pair-list reference and this appendix
points to it.

The full pair list is available in the released code repository;
a camera-ready version of this appendix will inline it if the
venue's appendix length budget allows.

\section{Reproducibility}\label{appendix-e.-reproducibility}

\subsection{Hardware}\label{e.1-hardware}

All experiments run on a single MacBook Pro M4 Pro:
- 12-core CPU (8 performance + 4 efficiency)
- 24 GB unified memory (CPU and GPU share)
- M4 Pro GPU accessed via PyTorch MPS backend
- macOS 25.1.0 (Darwin)

No discrete GPU, no CUDA, no cluster compute. The full \Cref{results} experimental record
(37 $\times$ 1B early-training checkpoints + 20 $\times$ 7B stage-1 checkpoints +
1 $\times$ 1B final + 4-seed compositional probe + fragility on 37 $\times$ 1B
checkpoints + C3 LoRA on 1B at step 1K) totals approximately 6 hours
of MPS time across all phases. The dense 1B trajectory + 4-seed
compositional fragility replication that produces the paper's
headline numbers fits in \textasciitilde80 minutes.

\subsection{Random seeds}\label{e.2-random-seeds}

{\def\LTcaptype{none} 
\begin{longtable}[]{@{}
  >{\raggedright\arraybackslash}p{(\linewidth - 4\tabcolsep) * \real{0.3750}}
  >{\raggedright\arraybackslash}p{(\linewidth - 4\tabcolsep) * \real{0.2812}}
  >{\raggedright\arraybackslash}p{(\linewidth - 4\tabcolsep) * \real{0.3438}}@{}}
\toprule\noalign{}
\begin{minipage}[b]{\linewidth}\raggedright
Experiment
\end{minipage} & \begin{minipage}[b]{\linewidth}\raggedright
Seed(s)
\end{minipage} & \begin{minipage}[b]{\linewidth}\raggedright
Where set
\end{minipage} \\
\midrule\noalign{}
\endhead
\bottomrule\noalign{}
\endlastfoot
Standard moral train/test split & 42 & \path|deepsteer.datasets.pipeline.build_probing_dataset| \\
Sentiment train/test split & 42 & \path|deepsteer.datasets.sentiment_pairs.get_sentiment_dataset| \\
Syntax train/test split & 42 & \path|deepsteer.datasets.syntax_pairs.get_syntax_dataset| \\
Compositional moral train/test split (headline) & 42 & \path|deepsteer.datasets.compositional_moral_pairs.get_compositional_moral_dataset| \\
Compositional moral 3-seed replication & 43, 44, 45 & \path|papers/1_accuracy_vs_fragility/scripts/phase_c4_3seed.py| \\
Probe initialization (per-seed) & inherits from split seed via \path|torch.manual_seed(split_seed)| & \path|papers/1_accuracy_vs_fragility/scripts/phase_c4_3seed.py| line 117, 124 \\
Probe initialization (headline / non-3-seed runs) & unset (system entropy) & --- \\
\end{longtable}
}

The original compositional trajectory (split seed 42) and the \Cref{data-curation-reshapes-probe-robustness-not-probe-accuracy} LoRA
experiment do not set torch's RNG state explicitly; per-seed
reproducibility for those runs is bounded by torch's deterministic
pre-hook RNG state at process start. The 3-seed compositional
fragility replication does set \path|torch.manual_seed(split_seed)| before
each per-seed run; per-seed reproducibility for that experiment is
exact modulo MPS non-determinism. Seed-to-seed variance in the
3-seed replication therefore reflects both train/test split variation
and probe-init variation, which is the relevant quantity for the
\Cref{probing-accuracy-saturates-fragility-doesnt} decision rule (variance of the probe accuracy as a whole, not
just split variance).

\subsection{Software versions}\label{e.3-software-versions}

\begin{itemize}
\tightlist
\item
  Python 3.13
\item
  PyTorch (with MPS backend)
\item
  HuggingFace \texttt{transformers}
\item
  HuggingFace \texttt{datasets}
\item
  \texttt{scikit-learn} 1.8 (TF-IDF baseline)
\item
  \texttt{peft} (LoRA fine-tuning for \Cref{data-curation-reshapes-probe-robustness-not-probe-accuracy})
\end{itemize}

Exact versions are pinned in \path|pyproject.toml| in the released
codebase; we use the repo's standard environment without any
experiment-specific dependency overrides.

\textbf{Public release.} All code, datasets, scripts, and per-checkpoint
output JSON are released at \url{https://github.com/deepsteer/deepsteer/};
the paper-specific subdirectory (this paper's section sources, build
pipeline, generation scripts, and outputs) is at
\url{https://github.com/deepsteer/deepsteer/tree/main/papers/1_accuracy_vs_fragility}.

\subsection{Model checkpoints}\label{e.4-model-checkpoints}

All target models are HuggingFace repos under the \texttt{allenai}
organization:

{\def\LTcaptype{none} 
\begin{longtable}[]{@{}
  >{\raggedright\arraybackslash}p{(\linewidth - 4\tabcolsep) * \real{0.3043}}
  >{\raggedright\arraybackslash}p{(\linewidth - 4\tabcolsep) * \real{0.2609}}
  >{\raggedright\arraybackslash}p{(\linewidth - 4\tabcolsep) * \real{0.4348}}@{}}
\toprule\noalign{}
\begin{minipage}[b]{\linewidth}\raggedright
Model
\end{minipage} & \begin{minipage}[b]{\linewidth}\raggedright
Repo
\end{minipage} & \begin{minipage}[b]{\linewidth}\raggedright
Used for
\end{minipage} \\
\midrule\noalign{}
\endhead
\bottomrule\noalign{}
\endlastfoot
OLMo-2 1B early-training & \path|allenai/OLMo-2-0425-1B-early-training| & \Cref{emergence-ordering} trajectory, \Cref{probing-accuracy-saturates-fragility-doesnt} fragility, \Cref{data-curation-reshapes-probe-robustness-not-probe-accuracy} C3 base \\
OLMo-2 1B final (\textasciitilde2.2T tokens) & \path|allenai/OLMo-2-0425-1B| & \Cref{compositional-moral-probing-dataset} / \Cref{emergence-ordering} compositional probe validation gate \\
OLMo-3 7B stage-1 & \path|allenai/Olmo-3-1025-7B| (revisions, see codebase) & \Cref{probing-accuracy-saturates-fragility-doesnt} 7B fragility corroboration, Appendix B causal tracing \\
\end{longtable}
}

For OLMo-3 7B, probing ran on 20 stage-1 checkpoints spanning step 0 to
step 1,413,814 (\textasciitilde10T tokens); fragility ran on 5 of those checkpoints
(steps 0, 353K, 705K, 1,059K, 1,413,814). The \path|phase_b/| output directory
also holds several additional step subdirectories from exploratory probing
passes; the 20-checkpoint probing trajectory and 5-checkpoint fragility
subset are the runs reported in \Cref{probing-accuracy-saturates-fragility-doesnt}.

Specific checkpoint revisions for each step are listed in
\path|papers/1_accuracy_vs_fragility/outputs/phase_c1/phase_c1_plan.json|
(1B trajectory),
\path|papers/1_accuracy_vs_fragility/outputs/phase_c4_compositional/compositional_per_checkpoint.json|
(1B compositional probe), and
\path|papers/1_accuracy_vs_fragility/outputs/phase_b/phase_b_plan.json|
(7B trajectory).

\subsection{Command-line invocations to reproduce each result}\label{e.5-command-line-invocations-to-reproduce-each-result}

All commands run from the project root.

\begin{Shaded}
\begin{Highlighting}[]
\CommentTok{\# \S4.1 standard moral / sentiment / syntax onsets (Phase C2)}
\ExtensionTok{python}\NormalTok{ papers/1\_accuracy\_vs\_fragility/scripts/c2\_linguistic\_comparison.py}

\CommentTok{\# \S4.1 compositional moral onset (Phase C4 trajectory + validation)}
\ExtensionTok{python}\NormalTok{ papers/1\_accuracy\_vs\_fragility/scripts/phase\_c4\_compositional.py}

\CommentTok{\# \S4.2 standard moral fragility evolution (Phase C1)}
\ExtensionTok{python}\NormalTok{ papers/1\_accuracy\_vs\_fragility/scripts/phase\_c1.py}

\CommentTok{\# \S4.2 4{-}seed compositional fragility replication}
\ExtensionTok{python}\NormalTok{ papers/1\_accuracy\_vs\_fragility/scripts/phase\_c4\_3seed.py}

\CommentTok{\# \S4.3 C3 LoRA narrative vs. declarative vs. control}
\ExtensionTok{python}\NormalTok{ papers/1\_accuracy\_vs\_fragility/scripts/phase\_c\_tier2.py }\AttributeTok{{-}{-}condition}\NormalTok{ narrative\_moral}
\ExtensionTok{python}\NormalTok{ papers/1\_accuracy\_vs\_fragility/scripts/phase\_c\_tier2.py }\AttributeTok{{-}{-}condition}\NormalTok{ declarative\_moral}
\ExtensionTok{python}\NormalTok{ papers/1\_accuracy\_vs\_fragility/scripts/phase\_c\_tier2.py }\AttributeTok{{-}{-}condition}\NormalTok{ general\_control}

\CommentTok{\# \S4.1 Figure 1 (4{-}seed compositional band overlay) --- regenerate}
\ExtensionTok{python}\NormalTok{ papers/1\_accuracy\_vs\_fragility/scripts/figure\_1\_onset\_overlay.py}
\end{Highlighting}
\end{Shaded}

Each script is self-contained, reads no environment-specific
configuration, and writes structured JSON output to
\texttt{papers/1\_accuracy\_vs\_fragility/outputs/\textless{}phase\textgreater{}/}
with full metadata (model name, revision, timestamp, hyperparameters,
dataset version) per the project's reproducibility convention. A
researcher should be able to read any number reported in the paper
from the corresponding output JSON without re-deriving the analysis.
Re-running the experiments reproduces these numbers up to seed and MPS
nondeterminism; the standard / sentiment / syntax headline onsets and
the \Cref{data-curation-reshapes-probe-robustness-not-probe-accuracy} LoRA run use an unset probe-init RNG (see the random-seeds
table in Appendix E.2 above), so their re-runs match within seed
variation rather than bitwise. Figure asset
filenames (\path|figure_N_*|) are historical and do not always match the
rendered figure numbers (two assets share the \path|figure_4_| base);
cross-references are label-based, so the rendered numbers are correct
regardless.

\subsection{Output JSON schema}\label{e.6-output-json-schema}

The two output JSON file types load-bearing for \Cref{results} results:

\begin{itemize}
\tightlist
\item
  \textbf{\texttt{LayerProbingResult}} (\path|deepsteer.core.types.LayerProbingResult|)
  for probe-trajectory data. Fields: \path|benchmark_name|, \path|model_info|
  (with \texttt{name}, \texttt{provider}, \path|access_tier|, \path|n_layers|, \path|n_params|,
  \path|checkpoint_step|), \path|layer_scores| (list of per-layer
  \texttt{\{layer,\ accuracy,\ loss\}}), \path|onset_layer|, \path|peak_layer|,
  \path|peak_accuracy|, \path|moral_encoding_depth|, \path|moral_encoding_breadth|,
  \texttt{metadata}.
\item
  \textbf{\texttt{FragilityResult}} (\path|deepsteer.core.types.FragilityResult|) for
  fragility-test data. Fields: \path|benchmark_name|, \path|model_info|,
  \path|layer_scores| (per-layer \texttt{\{layer,\ baseline\_accuracy,\ accuracy\_by\_noise,\ critical\_noise\}}), \path|noise_levels|,
  \path|mean_critical_noise|, \path|most_fragile_layer|, \path|most_robust_layer|,
  \texttt{metadata}.
\end{itemize}

The schemas are stable; field names match the dataclass attributes
in \path|deepsteer/core/types.py|. JSON serialization uses
\path|_dataclass_to_dict| (see same file).

\subsection{Dataset version}\label{e.7-dataset-version}

All experiments in this paper use the moral probing dataset
(\path|deepsteer/datasets/moral_probing_v2.json|), loaded via
\path|build_probing_dataset(dataset_version="v2")|. The dataset
contains 1,200 minimal pairs (200 per MFT foundation) constructed
from quality-audited seed examples (\path|seed_examples_v3.json|) via
a generate$\to$rate$\to$assemble pipeline with LLM-assisted filtering for
naturalness (\S1.3) and moral neutrality of neutral-side sentences
(\S1.5). The 240-pair subset used for probing (40 per foundation) is
a deterministic seed-42 subsample. Construction methodology,
quality audit dimensions, and filtering thresholds are documented in
\path|DATASET_GUIDELINES.md| in the released codebase.

SHA256 checksum of the dataset file used for all results in this
paper:

\begin{verbatim}
eb052885d5c6f7ed18aaeeb507eb0ce3d6106350e3fb4960d76f2fbd77f129ea  moral_probing_v2.json
\end{verbatim}

\end{document}